\documentclass[final,5p,times,twocolumn, sort&compress]{elsarticle}

\usepackage{amssymb}
\usepackage{amsmath}
\usepackage{hyperref}
\usepackage{hyperref}
\usepackage{graphicx,verbatim}
\usepackage{amsmath}
\usepackage{amssymb}
\usepackage{mathtools}
\usepackage{amsthm}
\usepackage[capitalize,noabbrev]{cleveref}
\usepackage{amsmath, amssymb}

\usepackage{multirow}
\usepackage{booktabs}
\usepackage[table,xcdraw]{xcolor} 
\usepackage{verbatim}
\usepackage{xcolor}
\definecolor{lightgreen}{RGB}{204, 255, 204}
\usepackage{amsmath}
\usepackage{subcaption}
\usepackage{lscape}    % For landscape orientation
\usepackage{amsthm}
\usepackage{booktabs}
\usepackage{enumitem}
\usepackage{algorithm}
\usepackage{algpseudocode}

\algnewcommand\Input{\item[\textbf{Input:}]}%
\algnewcommand\Output{\item[\textbf{Output:}]}%

\algblockdefx{Indent}{EndIndent}{}{}          % no printed text
\algrenewcommand\algorithmicindent{1.5em}     % (optional) indent size
\algnotext{Indent}\algnotext{EndIndent} 

\usepackage{natbib}

\newcommand{\bd}{\boldsymbol} %italic bold for vectors
\newcommand{\mb}{\mathbf} % upright bold for matrices
\newcommand{\be}{\begin{equation}}
\newcommand{\ee}{\end{equation}}
\usepackage{graphicx}  % Required for \resizebox
\usepackage{subcaption}

\definecolor{splatcolor}{RGB}{70,130,180}   % steel blue (soft blue)
\definecolor{blurcolor}{RGB}{60,179,113}    % medium sea green (soft green)
\definecolor{slicecolor}{RGB}{205,92,92}    % indian red (soft red)

\makeatletter
\renewcommand{\fnum@algorithm}{\fname@algorithm~\thealgorithm.}
\makeatother

\DeclareMathOperator{\SD}{SD}
\DeclareMathOperator{\LIS}{LIS}

\journal{arXiv}

\begin{document}

\begin{frontmatter}





\title{A False Discovery Rate Control Method Using a Fully Connected Hidden Markov Random Field for Neuroimaging Data}

%use optional labels to link authors explicitly to addresses:
\author[label1]{Taehyo Kim} 
\author[label1,label2]{Qiran Jia} 
\author[label3]{Mony J. de Leon} 
\author[label1]{Hai Shu\corref{cor1}} 

\author{\\
for the Alzheimer's Disease Neuroimaging Initiative\fnref{note1}
}

\affiliation[label1]{organization={Department of Biostatistics, School of Global Public Health, New York University},city={New York},state={NY},country={USA}}

\affiliation[label2]{organization={Department of Population and Public Health Sciences, University of Southern California},city={Los Angeles},state={CA},country={USA}}

\affiliation[label3]{organization={Brain Health
Imaging Institute, Department of Radiology,
Weill Cornell Medicine},city={New York},state={NY},country={USA}}

\cortext[cor1]{Corresponding author: hs120@nyu.edu}

\fntext[note1]{Data used in preparation of this article were obtained from the Alzheimer’s Disease
Neuroimaging Initiative (ADNI) database (adni.loni.usc.edu). As such, the investigators
within the ADNI contributed to the design and implementation of ADNI and/or provided data
but did not participate in analysis or writing of this report. A complete listing of ADNI
investigators can be found at:
\url{http://adni.loni.usc.edu/wp-content/uploads/how_to_apply/ADNI_Acknowledgement_List.pdf}.}

%% Abstract
\begin{abstract}
False discovery rate (FDR) control methods are essential for voxel-wise multiple testing in neuroimaging data analysis, where hundreds of thousands or even millions of tests are conducted to detect brain regions associated with disease-related changes. Classical FDR control methods (e.g., BH, q-value, and LocalFDR) assume independence among tests and often lead to high false non-discovery rates (FNR). Although various spatial FDR control methods have been developed to improve power, they still fall short of jointly addressing three major challenges in neuroimaging applications: capturing complex spatial dependencies, maintaining low variability in both false discovery proportion (FDP) and false non-discovery proportion (FNP) across replications, and achieving computational scalability for high-resolution data.
To address these challenges, we propose fcHMRF-LIS, a powerful, stable, and scalable spatial FDR control method for voxel-wise multiple testing. It integrates the local index of significance (LIS)-based testing procedure with a novel fully connected hidden Markov random field (fcHMRF) designed to model complex spatial structures using a parsimonious parameterization. We develop an efficient expectation-maximization algorithm incorporating mean-field approximation, the Conditional Random Fields as Recurrent Neural Networks (CRF-RNN) technique, and permutohedral lattice filtering, reducing the time complexity from quadratic to linear in the number of tests.
Extensive simulations demonstrate that fcHMRF-LIS achieves accurate FDR control, lower FNR, reduced variability in FDP and FNP, and a higher number of true positives compared to existing methods. Applied to an FDG-PET dataset from the Alzheimer's Disease Neuroimaging Initiative, fcHMRF-LIS identifies neurobiologically relevant brain regions 
and offers notable advantages in computational efficiency. 
%A Python package implementing  the  fcHMRF-LIS method is publicly available at \url{https://github.com/kimtae55/fcHMRF-LIS}.
\end{abstract}

%% Keywords

\end{frontmatter}



\section{Introduction}
\label{sec:intro}

Voxel-wise multiple testing is  indispensable 
in neuroimaging data analyses, which often involve
hundreds of thousands or even millions of 
voxel-level comparisons 
 to identify brain regions that exhibit statistically significant differences 
between population groups~\citep{worsley1996unified,Ashb00,Geno02,mirman2018corrections}. 
For example, 
in Alzheimer's disease (AD) research,
Fluorine-18 fluorodeoxyglucose positron emission tomography (FDG-PET) is widely used to measure brain glucose metabolism, serving as a neurodegeneration biomarker to support clinical diagnosis and track disease progression~\citep{ou2019fdg}. To uncover disease-related functional abnormalities, FDG-PET studies commonly conduct voxel-wise multiple testing across groups at different AD-related stages to localize brain regions of altered metabolic activity~\citep{Mosc05,Lee5,kantarci2021fdg}.

In such high-dimensional settings,
multiple testing methods 
 typically aim to control the false discovery rate (FDR), 
 a measure of type I error,
 defined as the mean of the false discovery proportion (FDP).
Compared with family-wise error rate control,
FDR control generally offers higher statistical power in large-scale multiple testing problems~\citep{Benj95}.
However, when applied to spatial data,
classical FDR control methods, such as  BH~\citep{Benj95}, q-value~\citep{storey2002direct}, and LocalFDR~\citep{efron2004large}, 
ignore spatial dependence
by treating tests as independent, which results in a substantial loss of statistical power~\citep{Shu15}. 
To mitigate this issue, 
various  spatial FDR control methods have been developed.
These include  random field-based approaches~\citep{Nguyen_McLachlan_Cherbuin_Janke_2014,Shu15, Sun15,kim2018peeling}, local smoothing-based procedures~\citep{zhang2011multiple,Tans18,laws,han2023spatially}, and deep learning-based methods~\citep{xia2017neuralfdr,kim2024deepfdr},
although not all were originally designed for neuroimaging data.
Despite these advances, current
spatial FDR control methods 
still fall short of jointly addressing
the following
three  major challenges 
 in neuroimaging applications.

First, existing spatial FDR control methods often fail to adequately capture the complex spatial dependencies inherent in neuroimaging data, such as 
spatial heterogeneity~\citep{Brod07}, 
distance-related dependence~\citep{perinelli2019dependence},
and long-range interactions~\citep{Liu13}.
Examples include  methods that 
assume
homogeneous {\it hidden Markov random fields} (HMRFs)~\citep{kim2018peeling} or
homogeneous Gaussian random fields~\citep{Sun15},
rely on (hidden) Markov random fields
with short-range neighborhoods~\citep{Nguyen_McLachlan_Cherbuin_Janke_2014,Shu15,kim2018peeling},
or employ
local smoothing approaches~\citep{zhang2011multiple,Tans18,laws,han2023spatially}.
This raises concerns about the validity of such methods in controlling the FDR or their optimality in minimizing the false non-discovery rate (FNR)~\citep{Geno02b}, 
 a measure of type II error,
defined as the mean of false non-discovery proportion (FNP).

\begin{figure*}[ht!] % Use [t] or [b] to control vertical placement
\centering
\includegraphics[width=\textwidth]{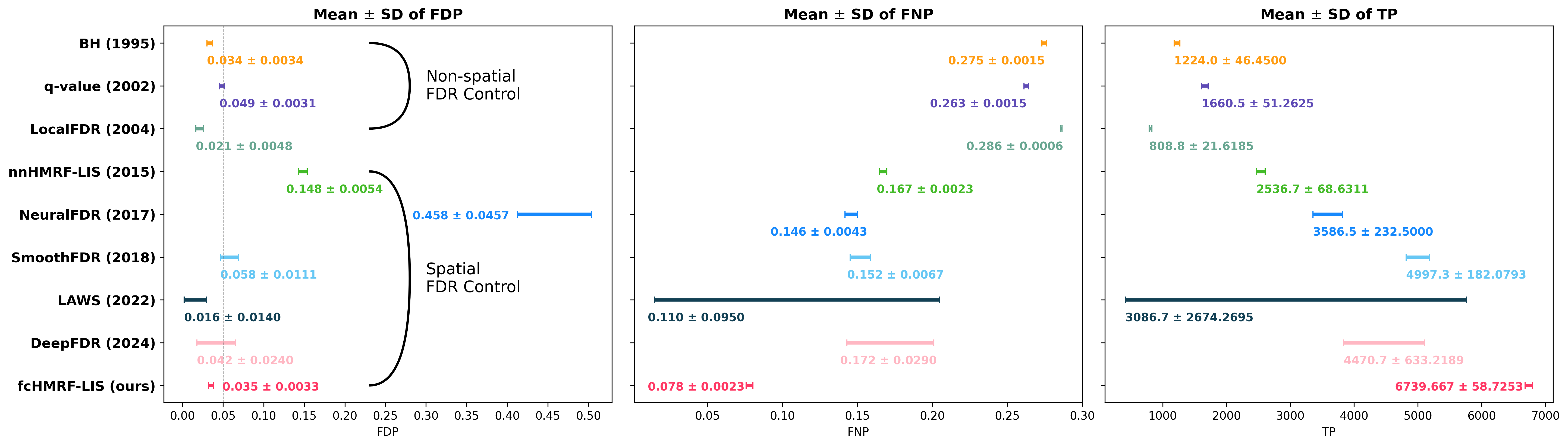}
    \caption{Comparison of nine FDR control methods under the  simulation setting in Section \ref{sim: setting} 
  with $(\mu_1,\sigma_1^2)=(-2, 1)$, approximately 30\% signal proportion, and a nominal FDR level of 0.05.}
    \label{fig:start}
\end{figure*}

Second, while
state-of-the-art methods focus on controlling the FDR
(the mean of FDP) and
some also aim to
minimize the FNR (the mean of FNP),  they often exhibit high variability  in  FDP or FNP, 
%{\color{red}true positives (TP)}
leading to instability  in results
across replications. Figure~\ref{fig:start} illustrates the variability in FDP, FNP, and  the number of true positives (TP) detected across 50 simulation replications, comparing nine FDR control methods: 
three non-spatial methods, BH~\citep{Benj95}, q-value~\citep{storey2002direct}, and LocalFDR~\citep{efron2004large};
nnHMRF-LIS~\citep{Shu15}, which uses a nearest-neighbor HMRF (nnHMRF);
two local smoothing-based methods, 
SmoothFDR~\citep{Tans18} and LAWS~\citep{laws}; 
two deep learning-based methods, 
NeuralFDR~\citep{xia2017neuralfdr} and DeepFDR~\citep{kim2024deepfdr};
and our proposed method.
Notably, although
the two most recent methods
LAWS and DeepFDR 
empirically control the
FDR at the nominal level of 0.05,
their standard deviations (SDs) of FDP
are large, exceeding three times that of BH,
and they also show unexpectedly high
SDs of FNP and TP.
This instability is particularly concerning in neuroimaging, where multiple testing is inherently exploratory and the truth status of each hypothesis is unobservable due to the unsupervised nature of the task.
If spatial FDR control methods yield inconsistent results across experiments, their reliability and practical utility  in neuroimaging applications are seriously undermined.

%Methods control FDP like LAWS, neuralFDR, 

Third, most spatial FDR control methods (e.g., \citep{Shu15,Sun15,xia2017neuralfdr,Tans18,laws,han2023spatially}) lack computational scalability, rendering them 
 computationally expensive
for analyzing large-scale, high-resolution neuroimaging datasets.
For example, in our real data analysis (Section~\ref{sec: real data})
using FDG-PET  data with 439,758 brain voxels of interest,
LAWS~\citep{laws} required an average of  149.5 hours on 20 CPU cores, while the deep learning-based method NeuralFDR~\citep{xia2017neuralfdr} still needed 
25.7 hours even when run on a GPU.

To address the three
challenges, 
we propose fcHMRF-LIS,
a novel spatial FDR control method for voxel-wise multiple testing in neuroimaging data,
which aims to control the FDR, minimize the FNR, and maintain low variability in both FDP and FNP.
The  fcHMRF-LIS method builds on the {\it local index of significance} (LIS)-based testing procedure~\citep{Sun09}, implemented under our proposed {\it fully connected} HMRF (fcHMRF) model.

The LIS-based testing procedure~\citep{Sun09}
is a  widely used 
framework for FDR control when tests are dependent.
The LIS
serves as a substitute for the traditional p-value.
For each test,
the LIS is defined as 
the conditional probability 
that the null hypothesis is true given 
the test statistics from all tests. 
Under mild conditions, the LIS-based
testing procedure 
asymptotically minimizes the FNR while controlling the FDR at a prespecified level~\citep{Sun09,xie2011optimal,Shu15}.
The key challenge in applying 
this procedure  lies in
accurately modeling the dependence among tests in order to estimate the unknown LIS values in practice. 
Previous works have explored this challenge in various data settings.
For one-dimensional data, such as influenza-like illness surveillance time series or single nucleotide polymorphisms (SNPs) on chromosomes in genome-wide association studies (GWAS), hidden Markov chain models have been used to estimate LIS values~\citep{Sun09,kim2024semi,Wei09}. In GWAS, other studies have employed HMRFs to model spatial dependence among SNPs using prior linkage disequilibrium information to define graph structures~\citep{li2010hidden,10.1214/16-AOAS956}.
In neuroimaging, nnHMRF-LIS~\citep{Shu15} and
its variant~\citep{kim2018peeling} use HMRFs with short-range neighborhoods, while DeepFDR~\citep{kim2024deepfdr} adopts 
a deep neural network called W-Net to learn complex spatial dependence.

However, as shown in Figure~\ref{fig:start},
neither nnHMRF-LIS nor DeepFDR perform well
in our simulation study. 
The failure of nnHMRF-LIS to control the FDR stems from its reliance on a short-range neighborhood structure, which is inadequate for capturing the complex spatial dependencies in neuroimaging data, the first major challenge outlined earlier. DeepFDR, on the other hand, suffers from substantial variability across replications, likely due to overfitting. Although it adopts a sophisticated W-Net architecture with millions of parameters, in the multiple testing setting, DeepFDR uses only a single image of test statistics and corresponding 
p-values as input, providing limited information for training.
In contrast, nnHMRF-LIS exhibits low variability, probably owing to its parsimonious model with only a few  parameters. This observation motivates our extension of nnHMRF to an fcHMRF model that retains a small number of parameters
 for low variability
 while effectively capturing
complex neuroimaging spatial dependencies.
 %incorporating the three key neuroimaging spatial dependencies: distance-related dependence, spatial heterogeneity, and long-range interactions.

Our proposed fcHMRF 
is inspired by
 the fully connected conditional random field (CRF) with  appearance and smoothness kernels, a powerful model widely used in image segmentation~\citep{Krähenbühl_Koltun_2012a,Zheng_Jayasumana_Romera2015,Arnab_Zheng_Jayasumana_Romera-Paredes_Larsson_Kirillov_Savchynskyy_Rother_Kahl_Torr_2018}.
In a fully connected CRF,
each voxel is represented  as a node in an undirected graph, and all pairs of nodes are connected by edges, 
which allows the model to capture long-range interactions.
The appearance and smoothness kernels further enable the model to account for spatial heterogeneity and distance-related dependence. 
We  innovatively formulate our fcHMRF 
as this fully connected CRF
to appropriately model the complex spatial dependencies in neuroimaging data.

Unlike standard fully connected CRF applications in supervised image segmentation, where large numbers of labeled images are available for training,
our multiple testing context is unsupervised. We do not observe the ground-truth hypothesis states and are limited to a single image of test statistics and their transformations.
To address this, we develop an expectation-maximization (EM) algorithm to estimate the parameters of the fcHMRF model and subsequently compute the LIS values for hypothesis inference.
Since  exact inference in the  fcHMRF is computationally intractable,
we employ the mean-field approximation method~\citep{koller2009probabilistic,Krähenbühl_Koltun_2012a} within both the EM algorithm and LIS computation.  
To implement the mean-field updates efficiently, we leverage  the CRFs-as-Recurrent-Neural-Networks (CRF-RNN) technique~\citep{Zheng_Jayasumana_Romera2015}, which unrolls mean-field iterations into a series of differentiable operations analogous to layers in a neural network. 
In addition, we incorporate the permutohedral lattice filtering technique~\citep{Adams_Baek_Davis_2010,baek_adams_proof} to accelerate the message-passing step within each mean-field iteration. 
Together, these techniques reduce the time complexity from quadratic to linear in the number of tests, enabling scalable inference for large neuroimaging datasets.
The resulting fcHMRF-LIS method thus combines spatial expressiveness, statistical stability, and computational efficiency.

Our main contributions are summarized as follows:
\begin{itemize}[leftmargin=1em]
    \item 
    We propose fcHMRF-LIS, a novel spatial FDR control method for voxel-wise multiple testing in neuroimaging data. The method  is built upon our newly designed fcHMRF, which 
   effectively captures complex spatial dependencies, including spatial heterogeneity, distance-related dependence, and long-range interactions, all of which are not jointly addressed by existing methods.

 \item The proposed fcHMRF-LIS integrates the LIS-based testing procedure~\citep{Sun09} to control the FDR while minimizing the FNR. 
 The underlying fcHMRF model is specifically designed with a small set of parameters, enabling fcHMRF-LIS 
 to maintain low variability in both FDP and FNP across replications, thereby offering improved stability over current deep learning and local smoothing-based methods.

    \item We develop an efficient EM algorithm for estimating the parameters of the fcHMRF model. To enable scalable inference, we incorporate the mean-field approximation~\citep{koller2009probabilistic,Krähenbühl_Koltun_2012a}, CRF-RNN technique~\citep{Zheng_Jayasumana_Romera2015}, and permutohedral lattice filtering~\citep{Adams_Baek_Davis_2010,baek_adams_proof}, reducing the time complexity from quadratic to linear in the number of tests.

  \item   We conduct extensive simulation studies to compare fcHMRF-LIS with eight existing FDR control methods~\citep{Benj95,storey2002direct,efron2004large,Shu15,xia2017neuralfdr,Tans18,laws,kim2024deepfdr}. 
  The results 
demonstrate the superiority of fcHMRF-LIS
in accurate FDR control, lower FNR, higher TP, and reduced variability across replications.
 
\item We apply fcHMRF-LIS to a large-scale FDG-PET dataset from the Alzheimer's Disease Neuroimaging Initiative (ADNI), analyzing 439,758 brain voxels to identify  regions associated with metabolic changes across AD-related stages. The fcHMRF-LIS yields neurobiologically relevant findings 
and achieves
significantly improved computational efficiency,
requiring only about 1.2 hours on a 20-core CPU server, with no GPU.
In contrast, the nnHMRF-LIS method~\citep{Shu15}
requires about 5.6 hours, even with GPU acceleration.

\end{itemize}

The rest of the paper is organized as follows.
Section~\ref{sec:method} introduces the proposed fcHMRF-LIS method and its  mean-field EM algorithm for parameter estimation.
Section~\ref{sec:simulation} compares fcHMRF-LIS
with eight existing FDR control methods via extensive simulations.
Section~\ref{sec: real data}
applies these nine methods
to analyze the FDG-PET data from ADNI. 
Section~\ref{sec:conclusion} concludes the paper.
Detailed implementation of the nine methods is provided in \ref{app sec: implement}.
A Python package implementing  the  fcHMRF-LIS method is publicly available at \url{https://github.com/kimtae55/fcHMRF-LIS}.

\section{Proposed Method}\label{sec:method}
\subsection{Problem Formulation}\label{sec:problem formulation}
Consider the problem of 
testing 
voxel-wise differences in 
population means 
between two comparison groups
based on 
a given measurement. 
In our motivating AD study,
to compare the brain glucose metabolism
between two groups,
for example, the AD group and the cognitively normal  group,
we test voxel-wise differences in their population means of 
the standardized uptake value ratio (SUVR) 
obtained from brain FDG-PET images.
 Assume  each subject has a 3D brain image with \( m \) voxels of interest. Let \( x_i \) be the test statistic for the null hypothesis \( \mathcal{H}_i \), which states no  difference in the mean SUVR values between the two groups at voxel~\( i \), i.e., \( \mu_{i1} - \mu_{i2} = 0\). Define  \( h_i \) as
a hidden state
representing the unobservable truth 
of \( \mathcal{H}_i \),
where
\( h_i = 1 \) if \( \mathcal{H}_i \) is false,
and \( h_i = 0 \) if 
\( \mathcal{H}_i \) is true. The goal of multiple testing is to infer the unknown hypothesis states \( \bd{h} = (h_1, \dots, h_m) \) from the observed test statistics \( \bd{x} = (x_1, \dots, x_m) \). 

 We aim to develop an optimal FDR control method
 for voxel-wise multiple testing,
 while also maintaining low variability
 in both 
 FDP and FNP. 
The FDR and FNR  
are the means of 
FDP and FNP, respectively,
formulated as
\begin{align*}
\text{FDR} = \mathbb{E}\big[\text{FDP}\big] \quad 
&\text{and}
\quad
\text{FNR} = \mathbb{E}\big[\text{FNP}\big], \quad\text{with}\\
\text{FDP} = \frac{N_{10}}{\max(N_1,1)} 
\quad 
&\text{and}
\quad
\text{FNP} = \frac{N_{01}}{\max(N_0,1)},
\end{align*}
where
the classification notation used for  tested hypotheses
is summarized in 
Table~\ref{table:c_matrix}.
Given a prespecified FDR level $\alpha$,
 an FDR control method is {\it valid} if it controls the FDR at level $\alpha$, 
 and is {\it optimal} if it minimizes the FNR among all valid
 FDR control methods. 
 For simplicity, false nulls and rejected nulls are referred to as {\it signals} and {\it discoveries}, respectively.

% TABLE 1: Classification Matrix
\begin{table}[h!]
\centering
\caption{Classification of tested hypotheses}
\label{table:c_matrix}
\begin{tabular}{c|ccc}
\hline
Number & Not rejected & Rejected & Total \\
\hline
True null & $N_{00}$ & $N_{10}$ & $m_0$\\
False null & $N_{01}$ & $N_{11}$ & $m_1$\\
Total & $N_0$ & $N_1$ & $m$\\
\hline
\end{tabular}
\end{table}

\subsection{LIS-based Testing Procedure}
The local index of significance (LIS) was proposed in \citep{Sun09} to replace the p-value for multiple testing under dependence. The LIS for hypothesis $\mathcal{H}_i$ is defined by 
\be\label{eqn: LIS}
\LIS_i(\bd{x}) = p(h_i = 0|\bd{x}),
\ee
which 
is the conditional probability 
of $h_i = 0$ given
\emph{all} test statistics $\bd{x} = (x_1, ...,x_m)$ not just the local statistic $x_i$. 
The LIS-based testing procedure for controlling the FDR at a prespecified level $\alpha$ is given as follows:
\be
\begin{split}\label{def: LIS procedure}
&\mbox{Let } k = \max\{j: \frac{1}{j}\sum_{i=1}^{j}\LIS_{(i)}(\bd{x}) \leq \alpha\}, \\
&\mbox{then reject all } \mathcal{H}_{(i)} 
\mbox{ with } i = 1,...,k.
\end{split}
\ee
Here, \( \LIS_{(i)}(\bd{x}) \) 
is the $i$-th smallest LIS value 
and $\mathcal{H}_{(i)}$ is 
the corresponding null hypothesis.
Due to the identity
$$
\text{FDR}=\mathbb{E}\left[\frac{1}{\max(R, 1)}\sum_{i=1}^R \LIS_{(i)}(\bd{x})\right],
$$ 
the LIS-based testing procedure in~\eqref{def: LIS procedure} is valid for controlling FDR at level $\alpha$.
Under mild conditions, 
it is also
 asymptotically optimal for minimizing the FNR~\citep{Sun09,xie2011optimal,Shu15}.

In practice, the true values  $\{\LIS_i(\bd{x})\}_{i=1}^m$
are unknown and must be
replaced by estimates
$\{\widehat{\LIS}_i(\bd{x})\}_{i=1}^m$
obtained from
a fitted model for the joint distribution 
$p(\bd{x},\bd{h})$.
The effectiveness of the LIS-based testing procedure thus hinges on how accurately the selected model captures the dependence  among tests.
To  model the complex spatial dependencies
present in neuroimaging data,
we propose an fcHMRF as the working model
for $p(\bd{x},\bd{h})$.
This fcHMRF generalizes the nearest-neighbor HMRF considered by \citep{Shu15}. Since  
the validity and optimality proofs in \citep{Shu15}
apply to a broad class of HMRF models, including our proposed fcHMRF,
the LIS-based 
testing procedure using the fcHMRF retains 
the validity and optimality.

\subsection{The fcHMRF Model}\label{sec: fcHMRF model}
We propose a novel fcHMRF to effectively capture 
the spatial dependence among tests.
The proposed fcHMRF assumes 
the observed  test statistics $\bd{x} = (x_1,\dots,x_m)$, 
are conditionally independent given the hidden states $\bd{h} = (h_1,\dots, h_m)$, 
leading to
the conditional probability density function:
\be\label{eqn: p(x|h)}
\begin{split}
p(\bd{x}|\bd{h})
  & = \prod_{i=1}^m p(x_i|h_i) \quad\text{with}\\
p(x_i|h_i)&=(1-h_i)f_0(x_i)+h_if_1(x_i),
\end{split}
\ee
where $f_0$ is a known null density function and $f_1$ is an unknown non-null density function.
In our context of testing the equality of two means, 
we define each $x_i$ as a z-statistic
which can be easily converted from a t-statistic \cite{Shu15}. Thus,
we set the null density $f_0$
to be the standard Gaussian density:
\be\label{eqn: f0=N}
f_0(x)=\varphi(x)=\frac{1}{\sqrt{2\pi}}\exp(-\frac{x^2}{2}).
\ee
The unknown non-null density $f_1$ will be estimated nonparametrically using kernel density estimation.

\iffalse
\begin{align*}
p(\bd{h})&=\frac{1}{Z}\exp\{-w_0\sum_{i=1}^m h_i -
\sum_{1\le i< j \le m} w_{ij} (h_i+h_j-2h_ih_j)\}\\
&=\frac{1}{Z}\exp\{\sum_{i=1}^m h_i(-w_0-\sum_{j\ne i}w_{ij})
+2\sum_{i<j}w_{ij} h_ih_j  \}\\
&=\frac{1}{Z}\exp\{\sum_{i=1}^m h_iw_i
+2\sum_{i<j}w_{ij} h_ih_j  \}\\
&=\frac{1}{Z}\exp\{\sum_{i=1}^m \sum_{j=1}^m w_{ij}h_ih_j
 \}
\end{align*}

\[
p(h_i,h_j)=C\exp(w_ih_i+w_jh_j + 2 w_{ij} h_ih_j + h_i  2\sum_{\ell \ne i,j}w_{i\ell} h_\ell + h_j  2\sum_{\ell \ne i,j}w_{j\ell} h_\ell )
\]
\fi

Our fcHMRF models the probability 
of hidden states $\bd{h}$ by a fully connected Markov random field as
\begin{align}\label{eqn: p(h)}
p(\bd{h})=
\frac{1}{Z}\exp\Big(-w_0\sum_{i=1}^m h_i-\sum_{1\le i<j\le m} w_{ij}|h_i-h_j|\Big),
\end{align}
where $Z$ is the normalizing constant, 
and
$w_0$ and $w_{ij}\ne 0$ are parameters for unary and pairwise potentials, respectively. In particular, $w_{ij}$  modulates the penalty for assigning different hidden state values to voxels $i, j$ based on their properties.
Since $w_{ij}\ne 0$ for all pairs of voxels,
the hidden states $h_i$ and $h_j$
of any two voxels
are conditionally dependent
given the hidden states of all other voxels.
Thus, every pair $(h_i,h_j)$
is connected in the undirected graph
representing the Markov random field $p(\bd{h})$~\citep{koller2009probabilistic}, making it a fully connected graph that enables long-range interactions.

We model $w_{ij}$ as
a combination of the appearance
and smoothness kernels:
\begin{align}
w_{ij} &= 
w_1 \underbrace{\exp(-\sum_{s \in S} \frac{|\ell_{i,s} - \ell_{j,s}|^2}{2\theta_{\alpha,s}^2} - \frac{|\Delta\mu_i - \Delta\mu_j|^2}{2\theta_\beta^2})}_{\text{\normalsize appearance kernel}} 
\label{eqn: wij}\\
&\qquad~~~~~+ w_2 \underbrace{\exp(-\sum_{s \in S} \frac{|\ell_{i,s} - \ell_{j,s}|^2}{2\theta_{\gamma,s}^2})}_{\text{\normalsize 
 smoothness kernel}}\nonumber.
\end{align}
where $\ell_{i,s}$, $s\in S=\{x,y,z\}$, are the 3D coordinates of voxel $i$, and $\Delta\mu_i$ is the difference in mean values at voxel $i$ 
between the two groups, which is estimated by $ \Delta\mu_i \approx \Delta\hat{\mu}_i =\bar{v}_{i1}-\bar{v}_{i2},$ with $\bar{v}_{ig}$ being the sample mean at voxel $i$ in group $g \in \{1,2\}$.
Note that in the appearance kernel,
the difference 
$\Delta\mu_i-\Delta\mu_j$ matters more than the individual values $\Delta\mu_i$ and $\Delta\mu_j$.

The above kernel combination is adapted  from the one used in fully connected CRFs in image segmentation~\citep{Krähenbühl_Koltun_2012a}. In our spatial multiple testing context, we replace the color intensity at voxel~$i$ used in their segmentation framework with the mean difference
$\Delta \mu_i$.
In image segmentation, 
the appearance kernel is inspired by the observation that nearby voxels with similar colors
($\Delta \mu_i$ in our context) 
are likely to have the same class (hypothesis state in our context),
and the smoothness kernel removes small isolated regions.
The degrees of nearness and similarity are controlled by
$\{\theta_{\alpha,s},\theta_{\gamma,s}\}_{s\in S}$ and $\theta_\beta$, respectively. The two kernels, together with the fully connected graphical model, can effectively capture spatial heterogeneity,  distance-related dependence, and long-range interactions~\citep{Krähenbühl_Koltun_2012a,Zheng_Jayasumana_Romera2015,Arnab_Zheng_Jayasumana_Romera-Paredes_Larsson_Kirillov_Savchynskyy_Rother_Kahl_Torr_2018}.

To simplify  the model parameter estimation,  
we fix $\{\theta_{\alpha,s},\theta_{\gamma,s}\}$ and $\theta_\beta$
to be the SDs of 
$\{\ell_{i,s}-\ell_{j,s}\}_{i\ne j}$
and $\{\Delta\mu_i-\Delta\mu_j\}_{i\ne j}$, respectively,
because
these $\theta$ parameters can be approximately absorbed into
the corresponding $w$ parameters. 
Specifically,
let $\theta\in\{ \theta_{\alpha,s}, \theta_{\gamma,s},\theta_\beta\}$, $\delta_{ij}\in \{\ell_{i,s}-\ell_{j,s}, \Delta\mu_i-\Delta\mu_j\}$,
and $w\in \{w_1,w_2\}$.
We use the ordinary least squares (OLS)
to approximate 
$\exp\{-\delta_{ij}^2/(2\theta^2)\}\in (0,1]$ as
\[
\exp\left(-\frac{\delta_{ij}^2}{2\theta^2}\right)
\approx b(\theta,\{\delta_{ij}\}_{i\ne j})\cdot
\exp\left(-\frac{\delta_{ij}^2}{2\SD^2(\{\delta_{ij}\}_{i\ne j})}\right).
\]
Then,
the OLS coefficient $b(\theta,\{\delta_{ij}\}_{i\ne j})$
can be absorbed into 
the corresponding learnable parameter $w$.
For  very large $m$, 
to avoid the computational burden of exactly computing
$\SD(\{\delta_{ij}\}_{i\ne j})$,
which requires calculating 
$\{\delta_{ij}\}_{i\ne j}$
with 
$O(m^2)$ time complexity, we instead estimate it 
using a large yet  computationally manageable subsample of $\{\delta_{ij}\}_{i\ne j}$.

From equations \eqref{eqn: p(x|h)}--\eqref{eqn: p(h)}, we obtain
the conditional probability of $\bd{h}$ given $\bd{x}$:
\be\label{eqn: p(h|x)}
p(\bd{h}|\bd{x}) \propto
\exp\Bigg\{
-\sum_{i=1}^{m} h_i \Big[w_0 + \log \frac{\varphi(x_i)}{f_1(x_i)} \Big]
	-	\sum_{1\le i<j\le m} w_{ij} |h_i - h_j|
\Bigg\},
\ee
which  also represents a fully connected Markov random field.

The parameter set of the fcHMRF
is 
$\bd{\phi}=\{f_1,\bd{w}=(w_0,w_1,w_2)\}$.
From the definition in \eqref{eqn: LIS},
$\LIS_i(\bd{x}) = p(h_i = 0|\bd{x};\bd{\phi})$
is the marginal probability that $h_i=0$ under the conditional distribution $p(\bd{h}|\bd{x};\bd{\phi})$.
Computing the exact marginal probability
$p(h_i = 0|\bd{x};\bd{\phi})$
is  intractable,
as it requires summing over
all $2^m$ possible configurations of $\bd{h}$
to obtain the normalizing constant 
in $p(\bd{h}|\bd{x};\bd{\phi})$,
followed by
$2^{m-1}$  addition operations to marginalize out 
$\bd{h}_{-i}$ (all entries of $\bd{h}$ except $h_i$).
One might consider approximating $\LIS_i(\bd{x})$  using  the Gibbs sampler by
$
\LIS_i(\bd{x}) 
\approx N^{-1}\sum_{n=1}^N \mathbb{I}(h_i^{(n,\bd{x})}=0)
$~\citep{Shu15},
where $\{\bd{h}^{(n,\bd{x})}=(h_1^{(n,\bd{x})},\dots,h_m^{(n,\bd{x})})\}_{n=1}^N$
are $N$ samples generated
by the Gibbs sampler~\citep{Gema84} from $p(\bd{h}|\bd{x};\bd{\phi})$,
and $\mathbb{I}(\cdot)$ is the indicator function.
However,
this approximation approach is  computationally expensive
due to the time complexity $O(m^2 (B+N))$
for 
the 
fully connected Markov random field
$p(\bd{h}|\bd{x};\bd{\phi})$, where 
$B$ is the number of burn-in iterations.
Instead, 
we approximate
$\LIS_i(\bd{x})$
by
\be\label{eqn: LIS mean-field}
\LIS_i(\bd{x})
= p(h_i = 0|\bd{x};\bd{\phi})\approx q_i(h_i=0|\bd{x};\bd{\phi}),
\ee
using
the mean-field approximation
$q(\bd{h}|\bd{x};\bd{\phi})=\prod_{i=1}^mq_i(h_i|\bd{x};\bd{\phi})$
to $p(\bd{h}|\bd{x};\bd{\phi})$~\citep{Krähenbühl_Koltun_2012a}.
The mean-field approximation  
is efficiently computed using the permutohedral lattice filtering method~\citep{Adams_Baek_Davis_2010,baek_adams_proof},
which reduces the time complexity to
$O(mR)$, where
$R$ is the number of mean-field iterations.
The details on
the mean-field approximation
and the permutohedral lattice filtering method
are provided in
Sections~\ref{sec: mean-field}
and~\ref{sec:permutohedral filtering}, respectively.

\subsection{The EM Algorithm for the Estimation of fcHMRF}\label{sec: EM alg}

In practice, 
the parameter set $\bd{\phi}$
of the fcHMRF is
unknown 
and is replaced
with its estimate 
when computing
the LIS values
using equation \eqref{eqn: LIS mean-field}.
We develop an
EM algorithm to estimate $\bd{\phi}$.

Let $\bd{\phi}^{(t+1)}$ denote the final estimate 
of 
$\bd{\phi}$ in the $(t+1)$-th iteration of the EM algorithm,
which aims to maximize
the following objective function
\begin{align}\label{eq:Q-function}
&Q(\boldsymbol{\phi}|\boldsymbol{\phi}^{(t)})
=
\sum_{\bd{h}} p(\bd{h}|\bd{x};\bd{\phi}^{(t)})
\log p(\bd{x},\bd{h};\bd{\phi})\\
&=\underbrace{\sum_{\bd{h}} 
    p(\bd{h}|\bd{x}; \boldsymbol{\phi}^{(t)})
    \log p\bigl(\bd{x}|\bd{h}; f_1\bigr)}_{\mbox{\normalsize  $Q_1(f_1 | \boldsymbol{\phi}^{(t)})$}}
  + \underbrace{\sum_{\bd{h}} 
    p\bigl(\bd{h}|\bd{x}; \boldsymbol{\phi}^{(t)}\bigr)
    \log p\bigl(\bd{h}; \boldsymbol{w}\bigr)}_{\mbox{\normalsize $
 Q_2(\boldsymbol{w} | \boldsymbol{\phi}^{(t)})$}}.\nonumber
\end{align}
We can maximize 
$Q(\boldsymbol{\phi}|\boldsymbol{\phi}^{(t)})$
for $\boldsymbol{\phi}$
by maximizing
$Q_1(f_1|\bd{\phi}^{(t)})$ for $f_1$
and $Q_2(\bd{w}|\bd{\phi}^{(t)})$ for $\bd{w}$,
separately.

However, 
similar to 
computing $\{\LIS_i(\bd{x})\}_{i=1}^m$ as mentioned at the end of Section~\ref{sec: fcHMRF model},
the exact maximization of 
$Q(\bd{\phi}|\bd{\phi}^{(t)})$
is intractable 
due to the 
normalizing constants 
in fully connected 
Markov random fields 
 $p(\bd{h}|\bd{x}; \boldsymbol{\phi}^{(t)})$
 and $p(\bd{h};\bd{w})$,
 and the approximate 
 maximization using  Gibbs-sampler
 samples from these fields
 is computationally expensive 
 because 
each update of $\bd{\phi}$ within the $(t+1)$-th   EM iteration requires
 $O(m^2(B+N))$ time~\citep{Shu15}.
 To address this,
 following \citep{zhang1992mean},
 we propose using
the mean-field approximations
of $p(\bd{h}|\bd{x}; \boldsymbol{\phi}^{(t)})$
 and $p(\bd{h};\bd{w})$
 in the EM algorithm,
but it still costs
 $O(m^2R+mN)$ time per update based on $N$ mean-field samples.
 We thus accelerate it using
 the permutohedral lattice filtering method~\citep{Adams_Baek_Davis_2010,baek_adams_proof},
reducing the time complexity to 
 $O(m(R+N))$ per update 
(see Section \ref{sec:permutohedral filtering}).

In $Q_1(f_1|\bd{\phi}^{(t)})$,
replacing $p(\bd{h}|\bd{x};\bd{\phi}^{(t)})$
with its mean-field approximation
$q(\bd{h}|\bd{x};\bd{\phi}^{(t)})=\prod_{i=1}^mq_i(h_i|\bd{x};\bd{\phi}^{(t)})$ yields
\begin{align}
&Q_1(f_1|\bd{\phi}^{(t)})
\approx
\hat{Q}_1(f_1|\bd{\phi}^{(t)})\nonumber\\
&=
\sum_{\bd{h}} q(\bd{h}|\bd{x};\bd{\phi}^{(t)})
\log p(\bd{x}|\bd{h};f_1)\nonumber\\
& 
=\sum_{\bd{h}} q(\bd{h}|\bd{x};\bd{\phi}^{(t)})
\sum_{i}[h_i\log f_1(x_i)+(1-h_i)\log \varphi(x_i)]\nonumber\\
&=
\sum_{i}\sum_{\bd{h}} [h_iq(\bd{h}|\bd{x};\bd{\phi}^{(t)})]\log f_1(x_i)\nonumber\\
&\qquad+
\sum_{i}\sum_{\bd{h}} [(1-h_i)q(\bd{h}|\bd{x};\bd{\phi}^{(t)})]
\log \varphi(x_i)
\nonumber\\
&=
\sum_{i}q_i(h_i=1|\bd{x};\bd{\phi}^{(t)})\log f_1(x_i)\nonumber\\
&\qquad+\sum_{i}q_i(h_i=0|\bd{x};\bd{\phi}^{(t)})\log \varphi(x_i).\label{Q1 approx}
\end{align}
%Note that the second term on the right hand side of \eqref{Q1 approx} is independent of $f_1$.

The optimal Gaussian-kernel estimator~\citep{kim2024semi}
of $f_1$ for maximizing $\hat{Q}_1(f_1|\bd{\phi}^{(t)})$  is
\begin{equation}\label{eq:f_1}
f_1^{(t+1)}(x)
=\frac{\sum_{i=1}^m q_i(h_i=1|\bd{x};\bd{\phi}^{(t)})
\frac{1}{h}\varphi(\frac{x-x_i}{h})}{\sum_{i=1}^m q_i(h_i=1|\bd{x};\bd{\phi}^{(t)})}.
\end{equation}
 We use the rule-of-thumb bandwidth \citep{silverman2018density}: $$h=0.9\cdot\min\left\{\SD(\{x_i\}_{i=1}^m),\frac{\text{IQR}(\{x_i\}_{i=1}^m)}{1.34}\right\}\cdot m_{\text{eff}}^{-1/5},$$
where $\text{IQR}(\{x_i\}_{i=1}^m)$ is the  interquartile range of $\{x_i\}_{i=1}^m$, and
the effective sample~size~\citep{Kish65} $$m_{\text{eff}}=\frac{[\sum_{i=1}^mq_i(h_i=1|\bd{x};\bd{\phi}^{(t)}) ]^2}{\sum_{i=1}^m [q_i(h_i=1|\bd{x};\bd{\phi}^{(t)}) ]^2}.$$

In $Q_2(\bd{w}|\bd{\phi}^{(t)})$,
replacing $p(\bd{h}|\bd{x};\bd{\phi}^{(t)})$
and $p(\bd{h};\bd{w})$
with their mean-field approximations
$q(\bd{h}|\bd{x};\bd{\phi}^{(t)})=\prod_{i=1}^m q_i(h_i|\bd{x};\bd{\phi}^{(t)})$
and $q(\bd{h};\bd{w})=\prod_{i=1}^m q_i(h_i;\bd{w})$, 
and then applying the Monte Carlo approximation,
we obtain
\begin{align}\label{eq:q2}
Q_2(\bd{w}|\bd{\phi}^{(t)})
&\approx
\sum_{\bd{h}}
q(\bd{h}|\bd{x};\bd{\phi}^{(t)})\log q(\bd{h};\bd{w}) \nonumber\\
&\approx
\hat{Q}_2(\bd{w}|\bd{\phi}^{(t)})=
\frac{1}{N}\sum_{n=1}^N \sum_{i=1}^m \log q_i(h_i= h_i^{(n,t)};\bd{w}),
\end{align}
where  $h_{i}^{(n,t)}$ is independently sampled  from the Bernoulli distribution $q_i(h_i|\bd{x};\bd{\phi}^{(t)})$ for all $n=1,\dots, N$ and $ i=1,\dots, m$.

The optimal $\bd{w}$
that minimizes 
 $-\hat{Q}_2(\bd{w}|\bd{\phi}^{(t)})$ %in~\eqref{eq:q2}
is obtained
using  
gradient descent,
where the gradient
is efficiently computed
via 
backpropagation~\citep{zhang2023dive},
a reverse-mode automatic differentiation method, 
through the CRF-RNN network~\citep{Zheng_Jayasumana_Romera2015}
that implements the mean-field approximation $\{q_i(h_i;\bd{w})\}_{i=1}^m$. 
The mean-field approximation
and its CRF-RNN implementation
are described in Section~\ref{sec: mean-field}.

\subsection{Mean-field Approximation and CRF-RNN} \label{sec: mean-field}
In the EM algorithm,
we replace the Markov random fields
$p(\bd{h}|\bd{x};\bd{\phi}^{(t)})$
and $p(\bd{h};\bd{w})$ 
in \eqref{eqn: p(h|x)} and \eqref{eqn: p(h)}
with their mean-field approximations
$q(\bd{h}|\bd{x};\bd{\phi}^{(t)})=\prod_{i=1}^m q_i(h_i|\bd{x};\bd{\phi}^{(t)})$
and $q(\bd{h};\bd{w})=\prod_{i=1}^m q_i(h_i;\bd{w})$.
We write 
$p(\bd{h}|\bd{x};\bd{\phi}^{(t)})$
and $p(\bd{h};\bd{w})$
in the form of CRFs,
which are widely used 
in image segmentation~\citep{Krähenbühl_Koltun_2012a,Zheng_Jayasumana_Romera2015,Arnab_Zheng_Jayasumana_Romera-Paredes_Larsson_Kirillov_Savchynskyy_Rother_Kahl_Torr_2018}:
\be\label{eqn: p(h|I)}
p(\bd{h}|\bd{I})\,{=}\,\frac{1}{Z(\bd{I})}
\exp\left\{ \sum_{i=1}^m U_i(h_i) 
\,{-}\,\sum_{1\le i<j\le m}|h_i\,{-}\,h_j|\sum_{l=1}^2w_l k_l(\bd{I}_i,\bd{I}_j)
\right\}
\ee
where image features
$\bd{I}=\{\bd{I}_i\}_{i=1}^m$  
with $\bd{I}_i=\{\{\ell_{i,s}\}_{s\in S},\Delta\mu_i\}$,
and $k_l(\cdot,\cdot)$ is a Gaussian kernel.
For simplicity, we omit the conditioning on $\bd{x}$, if applicable, in the notation $p(\bd{h}|\bd{I})$. We also drop the conditioning on $\bd{I}$ in $p(\bd{h}|\bd{I})$ throughout the rest of the paper.

The mean-field method~\citep{koller2009probabilistic}
finds a distribution 
$q(\bd{h})$,
which is close to $p(\bd{h})$
in the sense that it minimizes the 
Kullback–Leibler divergence
$$
D_{\text{KL}}(q(\bd{h})\| p(\bd{h}))
=\sum_{\bd{h}}q(\bd{h})\log\frac{q(\bd{h})}{p(\bd{h})}
$$
within the class of distributions
representable as a product of independent marginals:
\[
%p(\bd{h})\approx 
q(\bd{h})=\prod_{i=1}^mq_i(h_i).
\]
The mean-field approximation
$q(\bd{h})$ for $p(\bd{h})$ given in~\eqref{eqn: p(h|I)}
can be iteratively obtained by 
the following update equation~\citep{Krähenbühl_Koltun_2012a}:
\[
q_i(h)=\frac{1}{Z_i}
\exp\left\{
U_i(h)-\sum_{h'}|h-h'|\sum_{l=1}^2 w_l\sum_{j\ne i}k_l(\bd{I}_i,\bd{I}_j)q_j(h')
\right\}.
\]

\begin{figure}[htb!] % Use [t] or [b] to control vertical     \centering
    \includegraphics[width=0.5\textwidth]{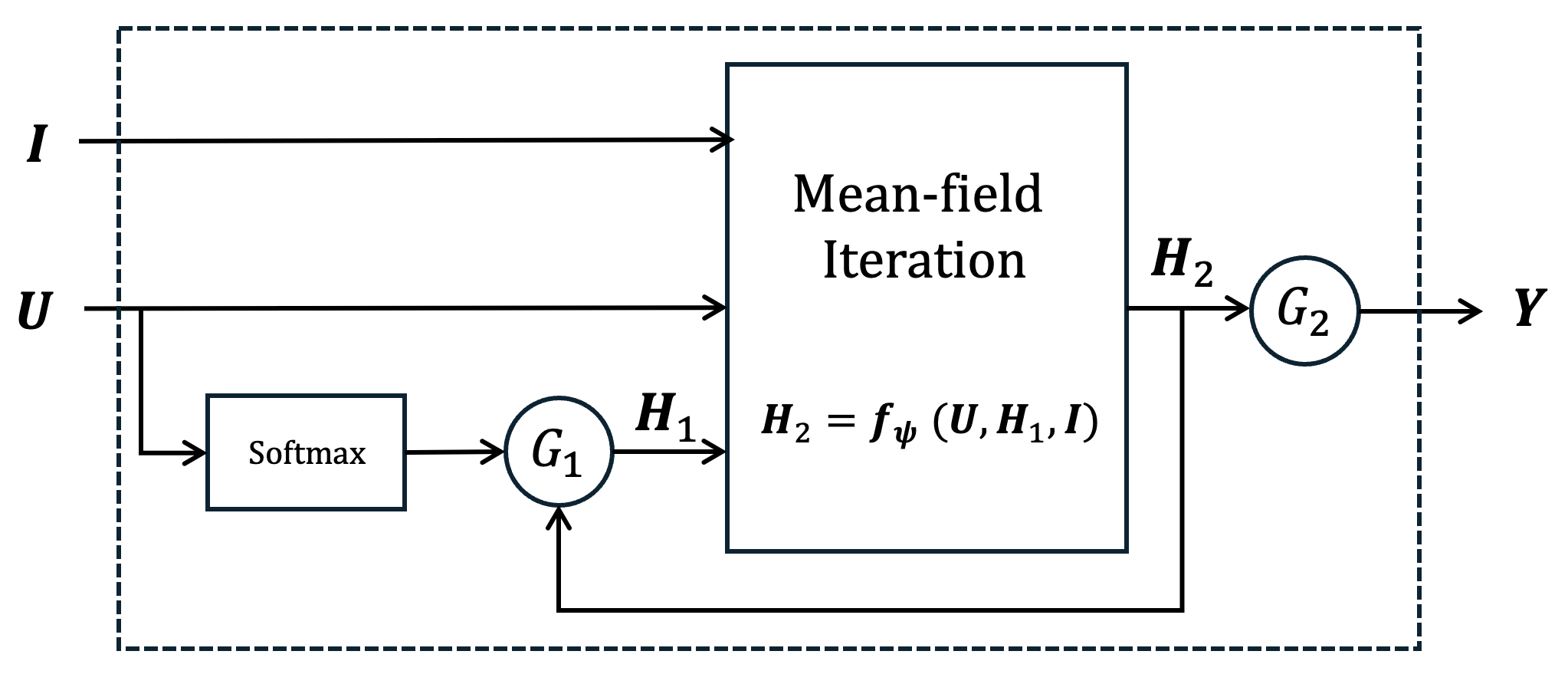}
    \caption{The CRF-RNN network. Gating functions $G_1$ and $G_2$ are given in \eqref{gate1} and \eqref{gate2}, respectively.}
    \label{fig: meanfield_as_rnn}
\end{figure}

The CRF-RNN method~\citep{Zheng_Jayasumana_Romera2015}
 decomposes
a single mean-field iteration
from the above update equation into simpler steps
that can be implemented using
neural network operations, as shown in Algorithm~\ref{alg:1}.
CRF-RNN  further formulates the
mean-field approximation algorithm, consisting of multiple mean-field iterations,
as an RNN~\citep{zhang2023dive}, a type of neural network,
where the output of one iteration is used as the input of the next iteration and all iterations share the same parameters. 
This makes implementing mean-field approximation simple and efficient by leveraging standard neural network libraries such as PyTorch~\citep{Pytorchpaper} and TensorFlow~\citep{TensorFlowpaper}. 
The CRF-RNN formulation of mean-field iterations is given by the following equations:
\begin{align}
\bd{H}_1(r) &= \begin{cases} 
\text{softmax}(\bd{U}), & r = 0, \\
\bd{H}_2(r-1), & 0 < r \leq R,
\end{cases} \label{gate1}\\
\bd{H}_2(r) &= \bd{f}_{\bd{\psi}}(\bd{U}, \bd{H}_1(r), \bd{I}), \quad 0 \leq r \leq R,  \\
\bd{Y}(r) &= \begin{cases} 
\bd{0}, & 0 \leq r < R, \\
\bd{H}_2(r), & r = R,
\end{cases} \label{gate2}
\end{align}
where $\bd{U}=\{U_i(h)\}_{i=1}^m$ is the set of voxel-wise unary potential values, $\bd{f}_{\bd{\psi}}(\bd{U}, \bd{q}_\text{in}, \bd{I})$ is the estimation of marginal distributions $\{q_i\}_{i=1}^m$ after one mean-field iteration, \( \bd{q}_\text{in} \) is the estimation from the previous iteration, and \( R \) is the number of iterations. 
The parameter set $\bd{\psi}$ is 
$\bd{\phi}^{(t)}$ for $p(\bd{h}|\bd{x};\bd{\phi}^{(t)})$
and is $\bd{w}$ for $p(\bd{h};\bd{w})$.
Figure~\ref{fig: meanfield_as_rnn} shows the CRF-RNN network.
The key advantage of CRF-RNN is that 
the parameter set $\bd{\psi}$ can be learned
using backpropagation~\citep{zhang2023dive}, 
a reverse-mode automatic differentiation method for neural networks, available in PyTorch and TensorFlow,
to efficiently compute the gradient 
for gradient descent.

    \begin{algorithm}[t!]
    \caption{Mean-field approximation}
    \label{alg:1}
    \begin{algorithmic}[1]
    \State $q_i(h) \gets \dfrac{\exp(U_i(h))}{\sum_{h'\in\{0,1\}}\exp(U_i(h'))},~i=1,\dots, m$ \Comment{Initialization}
    \Repeat
        \State $\tilde{q}_{i,l}(h) \gets \sum_{j \neq i} k_l(\bd{I}_i, \bd{I}_j) q_j(h),~i=1,\dots, m$  
      \Statex  \Comment{Message passing}
         \For{$i=1,\dots, m$}
        \State $\check{q}_i(h) \gets \sum_{l=1}^2 w_l \tilde{q}_{i,l}(h)$  
        \Comment{Weighting filter outputs}
        \State $\hat{q}_i(h) \gets \sum_{h' \in \{0,1\}} |h-h'| \check{q}_i(h')$ 
      \Statex   \Comment{Compatibility transform}
        \State $\breve{q}_i(h) \gets U_i(h) - \hat{q}_i(h)$ 
        \Comment{Add unary potentials}
     
        \State $q_i(h) \gets  \dfrac{\exp(\breve{q}_i(h))}{\sum_{h'\in\{0,1\}}\exp(\breve{q}_i(h'))}$  
      \Comment{Normalizing}
    \EndFor    
\Until{stopping criterion met}
    \end{algorithmic}
    \end{algorithm}

\subsection{Permutohedral Lattice Filtering}\label{sec:permutohedral filtering}
The message-passing step in
Algorithm \ref{alg:1}
for 
mean-field approximation
can be formulated as 
 Gaussian filtering:
\be\label{eqn: Gauss filter}
v_i'=\sum_{j=1}^m \exp\left(-\frac{1}{2}\| \bd{p}_i-\bd{p}_j\|_2^2\right) v_j,
~~i=1,\dots, m,
\ee
where the input value
$v_j=q_j(h)$,
the output value
$v_i'=\tilde{q}_{i,l}(h) +q_i(h)$,
and 
the position vector
$\bd{p}_i\in \mathbb{R}^d$
is 
$\big((\ell_{i,s}/\theta_{\alpha,s})_{s\in S}, \Delta\mu_i/\theta_\beta\big)^\top$
for $l=1$
and is 
$(\ell_{i,s}/\theta_{\gamma,s})_{s\in S}^\top$
for $l=2$
in our context.

The exact computation of this Gaussian filtering is computationally expensive, with time complexity $O(m^2d)$. 
To address this, the permutohedral lattice filtering method~\citep{Adams_Baek_Davis_2010,baek_adams_proof}
offers a highly  efficient 
approximation with
time complexity $O(md^2)$ and space complexity $O(md)$. 
This method
decomposes the filtering operation into three stages named {\it splatting}, {\it blurring} and {\it slicing},
which correspond to the familiar notions of
\textit{embedding}, \textit{convolution}, and \textit{reconstruction} in machine learning.

\iffalse
In the splatting stage, the input data is resampled onto the vertices of the permutohedral lattice; in the blurring stage, the permutohedral lattice undergoes linear convolution in order to simulate a Gaussian blur; in the slicing stage, the blurred data represented in the permutohedral lattice is resampled at the original locations, producing the output. In essence, the entire algorithm is a blurring operation flanked by two resampling operations. The purpose of the resampling operations is to modify the representation of the data in a way that can accommodate fast blurring.
\fi

Let
$H_d=\{\bd{s}\in\mathbb{R}^{d+1}| \bd{s}^\top \bd{1}_{d+1}=0\}$
be the $d$-dimensional hyperplane consisting of 
vectors in $\mathbb{R}^{d+1}$
whose coordinates sum to zero,
where $\bd{1}_{d+1}$ is the $(d+1)$-dimensional vector of all ones.
The {\it permutohedral lattice}
 in $H_d$, denoted by $A_d^*$,
is defined as
the projection of 
the scaled regular grid
$(d+1)\mathbb{Z}^{d+1}$
onto $H_d$.
The scale factor $d+1$ ensures that  lattice coordinates remain integer-valued.
Equivalently,
the permutohedral lattice $A_d^*$ can be expressed as the union of all remainder-$k$ points
in $H_d$ for $k=0,\dots, d$:
\[
A_d^*
=\bigcup_{k=0}^d \left\{
\bd{s}\in H_d\,|\,
\bd{s}=(d+1)\bd{z}+k\bd{1}_{d+1},
\bd{z}\in \mathbb{Z}^{d+1}
\right\}.
\]
The permutohedral lattice $A_d^*$
 tessellates the hyperplane $H_d$
 with uniform
simplices,  each of which is 
 a permutation and translation of a canonical simplex whose vertices are given by
\[
\bd{s}_k = (k\bd{1}_{d+1-k}^\top,-(d+1-k)\bd{1}_k^\top)^\top,
~~k=0,\dots, d,
\]
where each 
$\bd{s}_k$ is a remainder-$k$ point.

In the splatting stage,
the input values $\{v_i\}_{i=1}^m$
are embedded onto  the permutohedral lattice $A_d^*$.
Specifically, 
the position vector $\bd{p}_i$
of value $v_i$
is embedded into $H_d$
as
$\mb{E}\bd{p}_i$,
where $\mb{E}\in\mathbb{R}^{(d+1)\times d}$ has orthonormal columns spanning $H_d$:
\[
\mb{E} =
\left(
\begin{array}{cccc}
1 & 1 & \cdots & 1 \\
-1 & 1 & \cdots & 1 \\
0 & -2 & \cdots & 1 \\
\vdots & \vdots & \ddots & \vdots \\
0 & 0 & \cdots & -d
\end{array}
\right)
\left(
\begin{array}{cccc}
\frac{1}{\sqrt{2}} & 0 & \cdots & 0 \\
0 & \frac{1}{\sqrt{6}} & \cdots & 0 \\
\vdots & \vdots & \ddots & \vdots \\
0 & 0 & \cdots & \frac{1}{\sqrt{d(d+1)}}
\end{array}
\right).
\]
The embedded point 
$\mb{E}\bd{p}_i$
lies inside a simplex in $A_d^*$,
whose
vertices are denoted by 
$\{\bd{s}_{i,k}\}_{i=0}^d$.
Let $\bd{b}_i=(b_{i,0},\dots, b_{i,d})$
denote 
the barycentric coordinates of 
$\mb{E}\bd{p}_i$ within this simplex,
so that
\[
\mb{E}\bd{p}_i=\sum_{k=0}^d b_{i,k}\bd{s}_{i,k},
~\text{with~~$b_{i,k}\ge 0$~~and~~$\sum_{k=0}^d b_{i,k}=1$}.
\]
Then the value $v_i$ of point $\mb{E}\bd{p}_i$ is
distributed to each vertex
$\bd{s}_{i,k}$ with weight $b_{i,k}$,
i.e., 
$b_{i,k}v_i$ is accumulated at $\bd{s}_{i,k}$.

In the blurring stage,
the embedded values on the permutohedral lattice $A_d^*$ 
 are smoothed by a convolution that approximates Gaussian filtering. 
This is implemented by
convolving with the kernel $(1,2,1)$ along each of the $d+1$ lattice directions of the form
$\pm(1,\dots, 1, -d, 1,\dots,1)$.

The slicing stage mirrors the splatting stage in reverse.
Each  point 
$\{\mb{E}\bd{p}_i\}_{i=1}^m$
gathers the blurred values from the vertices of its enclosing simplex 
in lattice $A_d^*$
using the same barycentric weights 
$\bd{b}_i$, thereby reconstructing the filtered output $\{v_i'\}_{i=1}^m$.

The permutohedral lattice filtering method is summarized in Algorithm~\ref{alg:splat-blur-slice}.

\begin{algorithm}[t!]
\caption{Permutohedral lattice filtering}
\label{alg:splat-blur-slice}
\begin{algorithmic}[1]
\Input Input values $\{v_i\}_{i=1}^m$ and position vectors
$\{\bd{p}_i\}_{i=1}^m$
%\vspace{0.5em}
\State $V(\bd{s}) \gets 0, ~\forall \bd{s} \in A_d^*$ 
\For{$i=1,\dots, m$} \Comment{Splatting}
    \For{$k = 0,\dots,d$}
        \State $\bd{s} \gets$ the closest remainder-$k$ point in $A_d^*$ to $\mb{E}\bd{p}_i$ (by Theorem~4.10 in \cite{baek_adams_proof})  
        \State $b \gets$ the barycentric coordinate 
        of $\mb{E}\bd{p}_i$ at $\bd{s}$
        (by Proposition 5.2 in \cite{baek_adams_proof})
        \State $V(\bd{s}) \gets V(\bd{s}) + b v_i$
    \EndFor
\EndFor
%\vspace{0.5em}
\For{$k = 0,\dots, d$} \Comment{Blurring}
    \State $\bd{t} \gets 
    (\bd{1}_{k}^\top, -d, \bd{1}_{d-k}^\top)^\top$
    \For{all $\bd{s} \in A_d^*$}
        \State $W(\bd{s}) \gets \dfrac{V(\bd{s} - \bd{t})}{4} + \dfrac{V(\bd{s})}{2} + \dfrac{V(\bd{s} + \bd{t})}{4}$    \EndFor
    \State $V \gets W$
\EndFor
%\vspace{0.5em}
\For{$i=1,\dots, m$} \Comment{Slicing}
\State $v_i'\gets 0$
    \For{$k = 0,\dots, d$}
        \State $\bd{s} \gets$ the closest remainder-$k$ point in $A_d^*$
        to $\mb{E}\bd{p}_i$
        \State $b \gets$ the barycentric coordinate
        of $\mb{E}\bd{p}_i$ at $\bd{s}$
        \State $v_i' \gets v_i' + b W(\bd{s})$
    \EndFor
\EndFor
\Output $\{v_i'\}_{i=1}^m$
\end{algorithmic}
\end{algorithm}

\subsection{The Complete fcHMRF-LIS Algorithm}\label{sec: complete_fchmrf}

From equation \eqref{eqn: LIS mean-field}, 
the LIS estimates are given by $\widehat{\text{LIS}}_i(\bd{x})=q_i(h_i=0|\bd{x};\hat{\bd{\phi}})$ for $i=1,\dots, m$,
where 
$\prod_{i=1}^m q_i(h_i|\bd{x};\hat{\bd{\phi}})$
is the mean-field approximation
of $p(\bd{h}|\bd{x};\hat{\bd{\phi}})$
under the proposed fcHMRF model in Section~\ref{sec: fcHMRF model},
with $\hat{\bd{\phi}}$ denoting the 
estimate of the model parameter set
$\bd{\phi}=\{f_1,\bd{w}\}$.
To obtain $\hat{\bd{\phi}}$, we apply the mean-field  EM algorithm described in Section~\ref{sec: EM alg}. In the $(t+1)$-th EM iteration, the non-null density estimate $f_1^{(t+1)}$ is computed using equation~\eqref{eq:f_1}, and the 
weight estimate
$\bd{w}^{(t+1)}$
is obtained by  minimizing $-\hat{Q}_2(\bd{w}|\bd{\phi}^{(t)})$  in \eqref{eq:q2} via gradient descent. We implement gradient descent using the AdamW optimizer~\citep{adamW}, with the gradient of $-\hat{Q}_2(\bd{w}|\bd{\phi}^{(t)})$ is automatically computed via backpropagation~\citep{zhang2023dive}. 
The mean field approximations $\{q_i(h_i|\bd{x};\bd{\phi}^{(t)})\}_{i=1}^m$     and $\{q_i(h_i;\bd{w})\}_{i=1}^m$,  used in~\eqref{eq:f_1} and~\eqref{eq:q2}, 
are computed using the mean-field algorithm in Algorithm~\ref{alg:1}, efficiently implemented via CRF-RNN \eqref{gate1}-\eqref{gate2}.  The message passing step in Algorithm~\ref{alg:1} is  approximated using the permutohedral lattice filtering method described  in Algorithm~\ref{alg:splat-blur-slice}. 
We independently sample  $h_{i}^{(n,t)}$  from the Bernoulli distribution $q_i(h_i|\bd{x};\bd{\phi}^{(t)})$ for all $n=1,\dots, N$ and $ i=1,\dots, m$,
and use these samples to compute $-\hat{Q}_2(\bd{w}|\bd{\phi}^{(t)})$ 
and its gradient. 
convergence of the EM algorithm is monitored using the approximate 
$Q$-function,
$\hat{Q}(\bd{\phi}^{(t+1)}|\bd{\phi}^{(t)}) = \hat{Q}_1(f_1^{(t+1)}|\bd{\phi}^{(t)})+\hat{Q}_2(\bd{w}^{(t+1)}|\bd{\phi}^{(t)})$,
where $\hat{Q}_1$ and $\hat{Q}_2$
are defined in~\eqref{Q1 approx} and~\eqref{eq:q2}, respectively. 
%EM iterations, CRF-RNN iterations (5 iterations), AdamW iterations (5 epochs, we only have 1 sample). 
After obtaining  the final estimate $\hat{\bd{\phi}}$ from the EM algorithm, the  LIS estimates,
$\widehat{\text{LIS}}_i(\bd{x})=q_i(h_i=0|\bd{x};\hat{\bd{\phi}}), i=1,\dots, m$, are computed via the same mean-field algorithm based on CRF-RNN and permutohedral lattice filtering.
The resulting estimates $\{\widehat{\text{LIS}}_i(\bd{x})\}_{i=1}^m$ are then plugged into the LIS-based testing procedure~\eqref{def: LIS procedure} to obtain the multiple testing results. 
Algorithm~\ref{alg:fcHMRF-LIS algo} summarizes the steps of the proposed fcHMRF-LIS algorithm. 
Details of the implementation used in our simulations and real-data analysis are  provided in \ref{app sec: implement}.

\begin{algorithm}[h!]
\caption{The fcHMRF-LIS algorithm}
\label{alg:fcHMRF-LIS algo}
\begin{algorithmic}[1]
\Input A 3D volume of z-statistics $\bd{x}$ and 
 corresponding 3D volume of mean-difference estimates $\{\Delta\hat{\mu}_i\}_{i=1}^m$.

\State Initialize $\bd{\phi}^{(0)}$.
%\State $h_i \leftarrow 1 - 2 \left(1 - \Phi\left( \left| x_i \right| \right)\right)$ \hfill (Initialization)
\Repeat ~$t = 0,1,\dots$   \Comment{EM algorithm}
 \State  Compute $\{q_i(h_i|\bd{x};\bd{\phi}^{(t)})\}_{i=1}^m$
via mean-field iterations (Algorithm~\ref{alg:1}) implemented 
by CRF-RNN \eqref{gate1}-\eqref{gate2},
where message passing \eqref{eqn: Gauss filter}
is approximated by permutohedral lattice filtering
(Algorithm~\ref{alg:splat-blur-slice}). \label{mean-field step}
    \State Compute $f_1^{(t+1)}$ from~\eqref{eq:f_1}
    using $\{q_i(h_i|\bd{x};\bd{\phi}^{(t)})\}_{i=1}^m$.
    \State Compute $\bd{w}^{(t+1)}$ by minimizing
$-\hat{Q}_2(\bd{w}|\bd{\phi}^{(t)})$  in~\eqref{eq:q2}:
\Indent

    \State   Sample 
$h_i^{(n,t)}$
independently from $q_i(h_i|\bd{x};\bd{\phi}^{(t)})$
for all $n=1,\dots,N$ and
$i=1,\dots, m$;
 \Repeat   \Comment{Gradient descent}
        \State
        Compute
$\{q_i(h_i;\bd{w})\}_{i=1}^m$ 
as in Step \ref{mean-field step};

        \State Compute the gradient of
        $-\hat{Q}_2(\bd{w}|\bd{\phi}^{(t)})$
        via backpropagation
        using $\{q_i(h_i;\bd{w})\}_{i=1}^m$ 
        and $\{\{h_i^{(n,t)}\}_{n=1}^N\}_{i=1}^{m}$;
\State 
Update $\bd{w}$ using a gradient descent step;

  \Until stopping criterion met.
\EndIndent

\State $\bd{\phi}^{(t+1)}\gets \{f_1^{(t+1)},\bd{w}^{(t+1)}\}$.

\State $\hat{Q}(\bd{\phi}^{(t+1)}|\bd{\phi}^{(t)}) \gets \hat{Q}_1(f_1^{(t+1)} | \boldsymbol{\phi}^{(t)}) + \hat{Q}_2(\boldsymbol{w}^{(t+1)} | \boldsymbol{\phi}^{(t)})$.
\Until stopping criterion met.
\State Compute $\{q_i(h_i=0|\bd{x};\bd{\phi}^{(t+1)})\}_{i=1}^m$ as in 
 Step \ref{mean-field step}.

\State $\widehat{\text{LIS}}_i(\bd{x}) \gets q_i(h_i=0|\bd{x};\bd{\phi}^{(t+1)})$, $i=1,\dots,m$.

\State Run the LIS-based testing procedure~\eqref{def: LIS procedure} with $\{\widehat{\text{LIS}}_i(\bd{x})\}_{i=1}^m$.

\Output A 3D volume of estimates for the null hypothesis states~$\bd{h}$.
\end{algorithmic}
\end{algorithm}

\section{Simulation Studies}\label{sec:simulation}

\subsection{Methods for Comparison}\label{sec: method compare}

We conduct extensive simulations to
compare the proposed fcHMRF-LIS method with 
eight existing
FDR control methods, including BH~\citep{Benj95}, q-value~\citep{storey2002direct}, LocalFDR~\citep{efron2004large}, nnHMRF-LIS~\citep{Shu15}, NeuralFDR~\citep{xia2017neuralfdr}, SmoothFDR~\citep{Tans18}, LAWS~\citep{laws}, and DeepFDR~\citep{kim2024deepfdr}. 
The BH, q-value, and LocalFDR methods are classic FDR control methods developed for independent tests, while nnHMRF-LIS, NeuralFDR, SmoothFDR, LAWS, and DeepFDR are spatial methods applicable to 3D image data.

% 0.05 threshold
\begin{figure*}[h!]
  \centering
  \begin{subfigure}[t]{0.49\textwidth}
    \includegraphics[width=\textwidth]{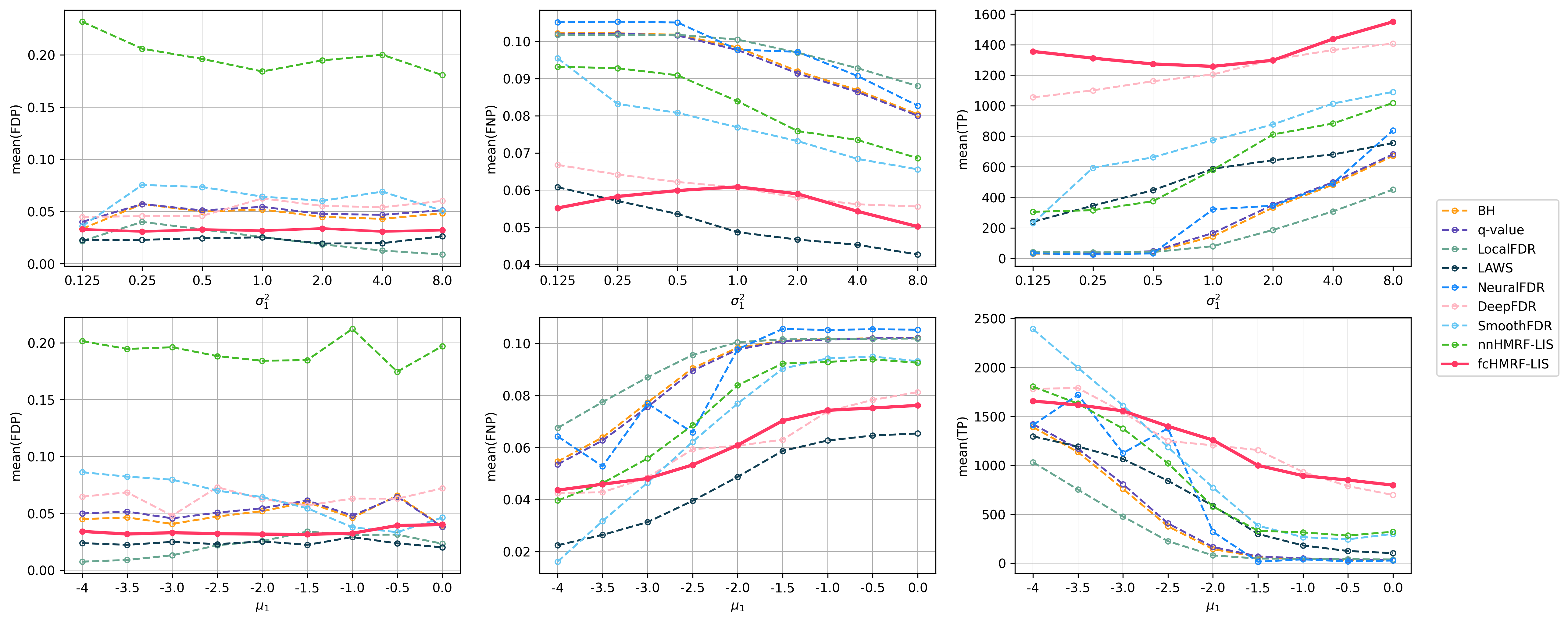}
    \caption{The means of FDP, FNP, and TP for 
    the cube with about $10\%$ signal proportion.
 The mean FDP values of NeuralFDR are omitted because they exceed 0.64.}
    \label{fig: sim 0.05 p=0.1 mean}
  \end{subfigure}
  \hfill
  \begin{subfigure}[t]{0.49\textwidth}
    \includegraphics[width=\textwidth]{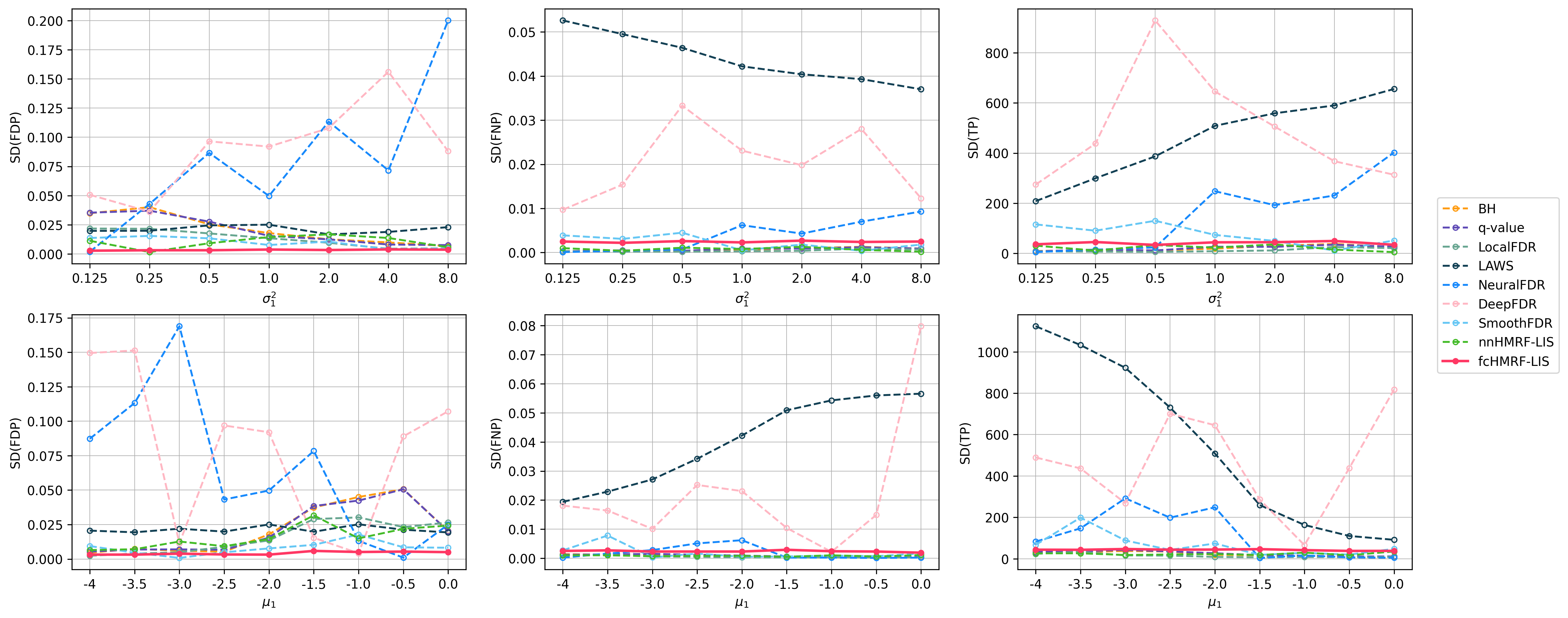}
    \caption{The SDs of FDP, FNP, and TP for 
    the cube with about $10\%$ signal proportion.}
    \label{fig: sim 0.05 p=0.1 std}
  \end{subfigure}

\vskip\baselineskip

  \begin{subfigure}[t]{0.49\textwidth}
    \includegraphics[width=\textwidth]{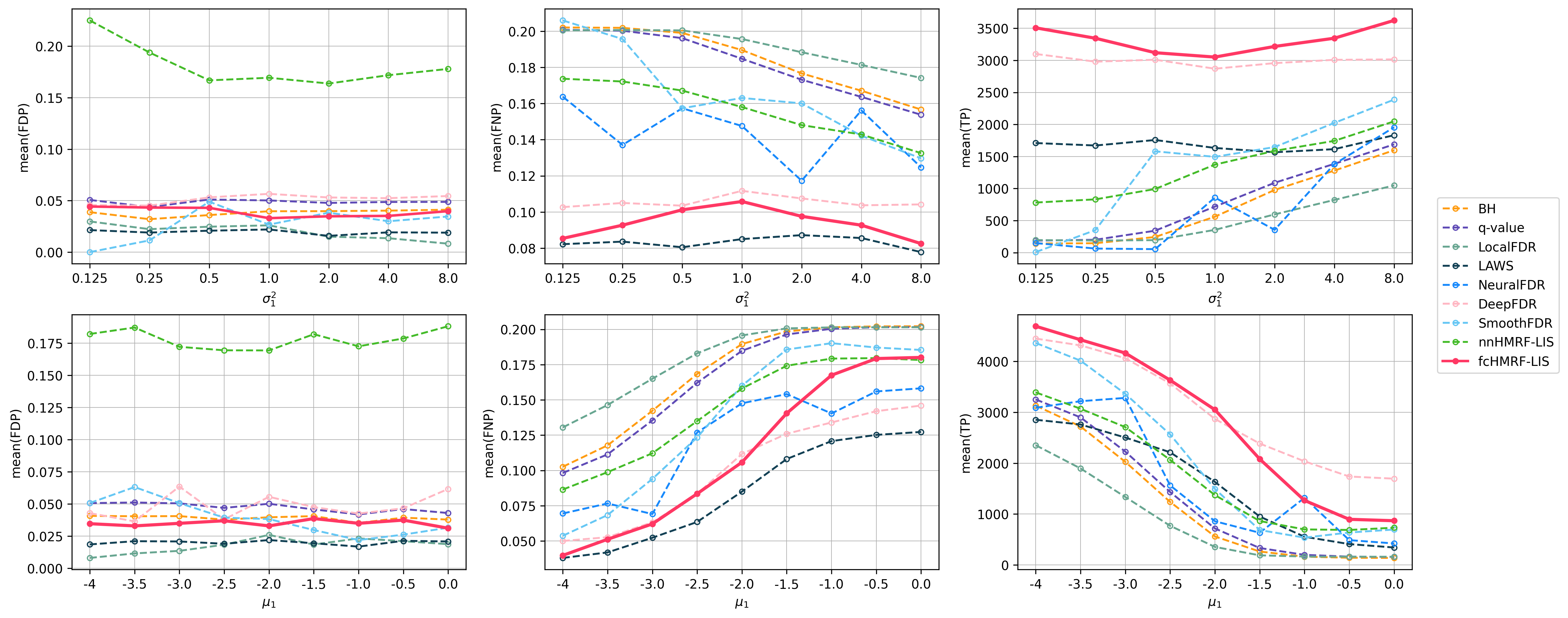}
    \caption{The means of FDP, FNP, and TP for 
    the cube with about $20\%$ signal proportion.
 The mean FDP values of NeuralFDR are omitted because they exceed 0.51.}
    \label{fig: sim 0.05 p=0.2 mean}
  \end{subfigure}
  \hfill
    \begin{subfigure}[t]{0.49\textwidth}
    \includegraphics[width=\textwidth]{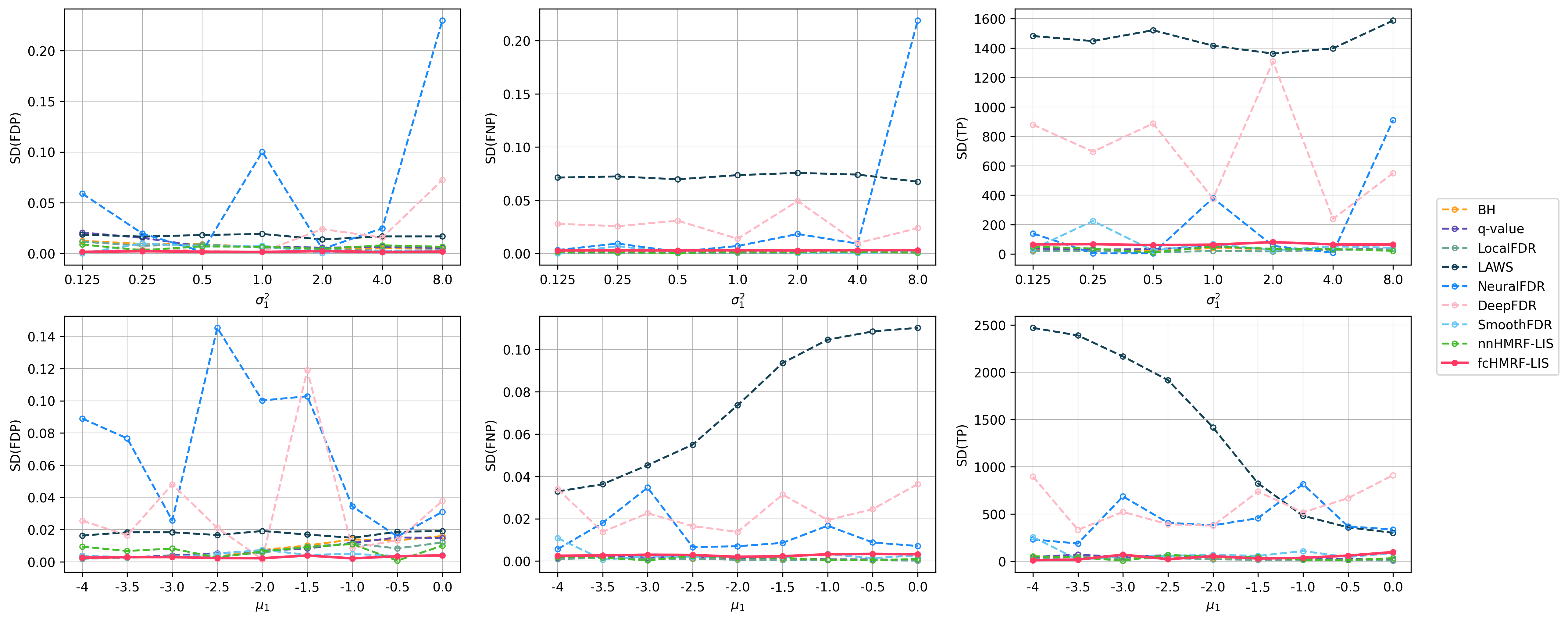}
    \caption{The SDs of FDP, FNP, and TP for 
    the cube with about $20\%$ signal proportion.}
    \label{fig: sim 0.05 p=0.2 std}
  \end{subfigure}

\vskip\baselineskip

  \begin{subfigure}[t]{0.49\textwidth}
    \includegraphics[width=\textwidth]{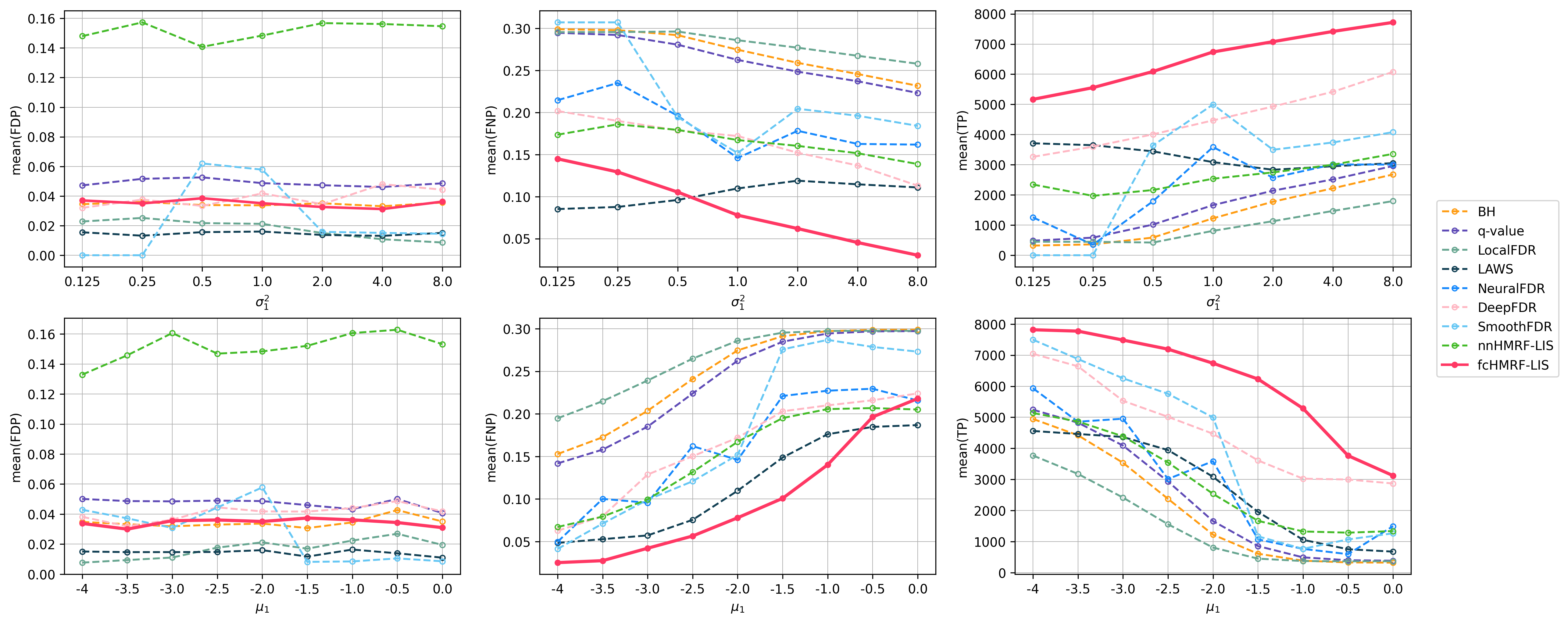}
    \caption{The means of FDP, FNP, and TP for 
    the cube with about $30\%$ signal proportion.
 The mean FDP values of NeuralFDR are omitted because they exceed 0.45.}
    \label{fig: sim 0.05 p=0.3 mean}
  \end{subfigure}
  \hfill
  \begin{subfigure}[t]{0.49\textwidth}
    \includegraphics[width=\textwidth]{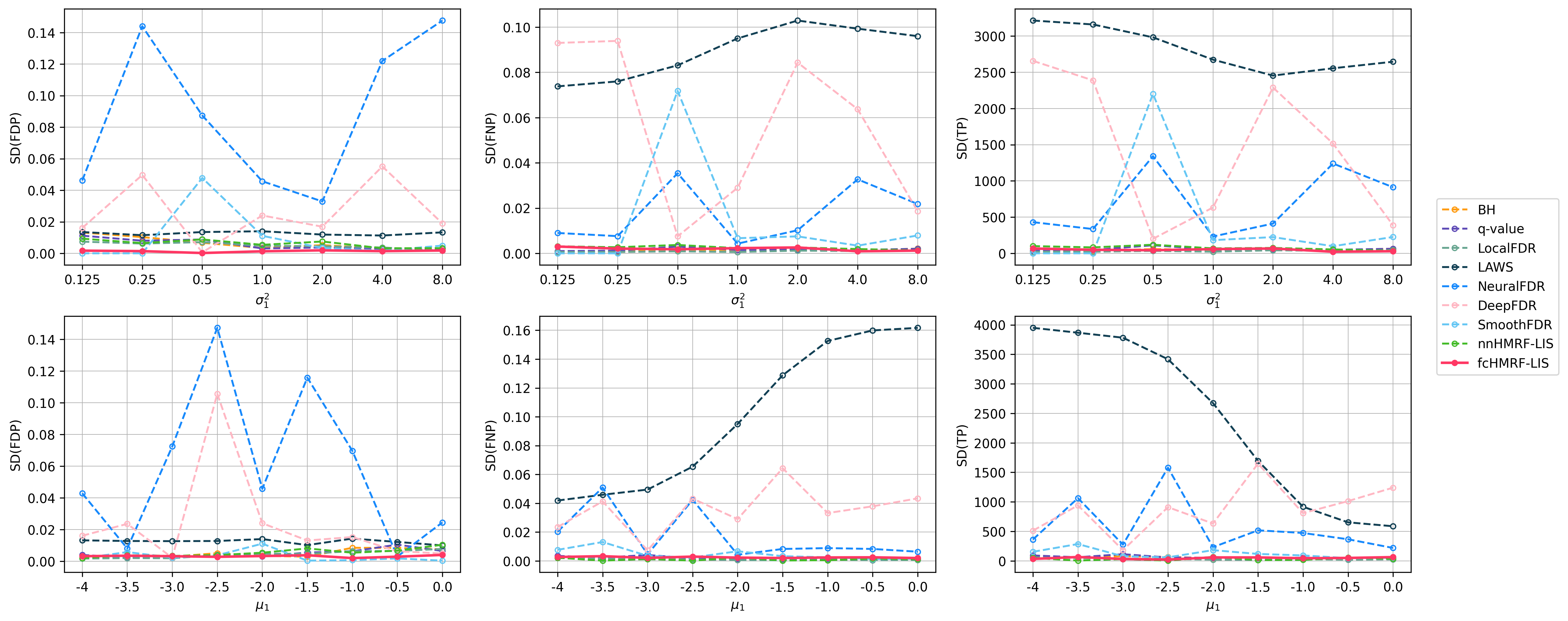}
    \caption{The SDs of FDP, FNP, and TP for 
    the cube with about $30\%$ signal proportion.}
    \label{fig: sim 0.05 p=0.3 std}
  \end{subfigure}
  
    \caption{Simulation results over 50 replications
    at a nominal FDR level of 0.05.}
    \label{fig: sim 0.05}
\end{figure*}

% 0.01 threshold
\begin{figure*}[h!]
  \centering
  \begin{subfigure}[t]{0.49\textwidth}
    \includegraphics[width=\textwidth]{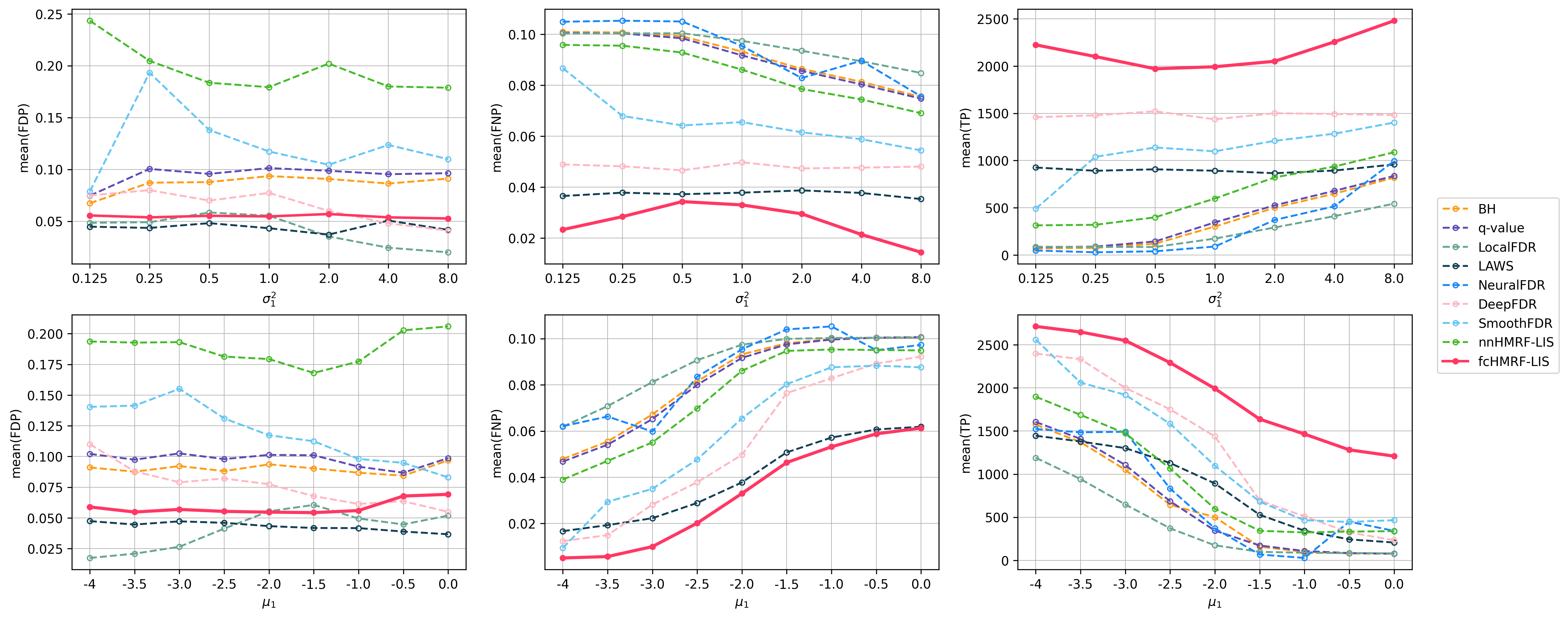}
    \caption{The means of FDP, FNP, and TP for 
    the cube with about $10\%$ signal proportion.
 The mean FDP values of NeuralFDR are omitted because they exceed 0.58.}
    \label{fig: sim 0.1 p=0.1 mean}
  \end{subfigure}
  \hfill
  \begin{subfigure}[t]{0.49\textwidth}
    \includegraphics[width=\textwidth]{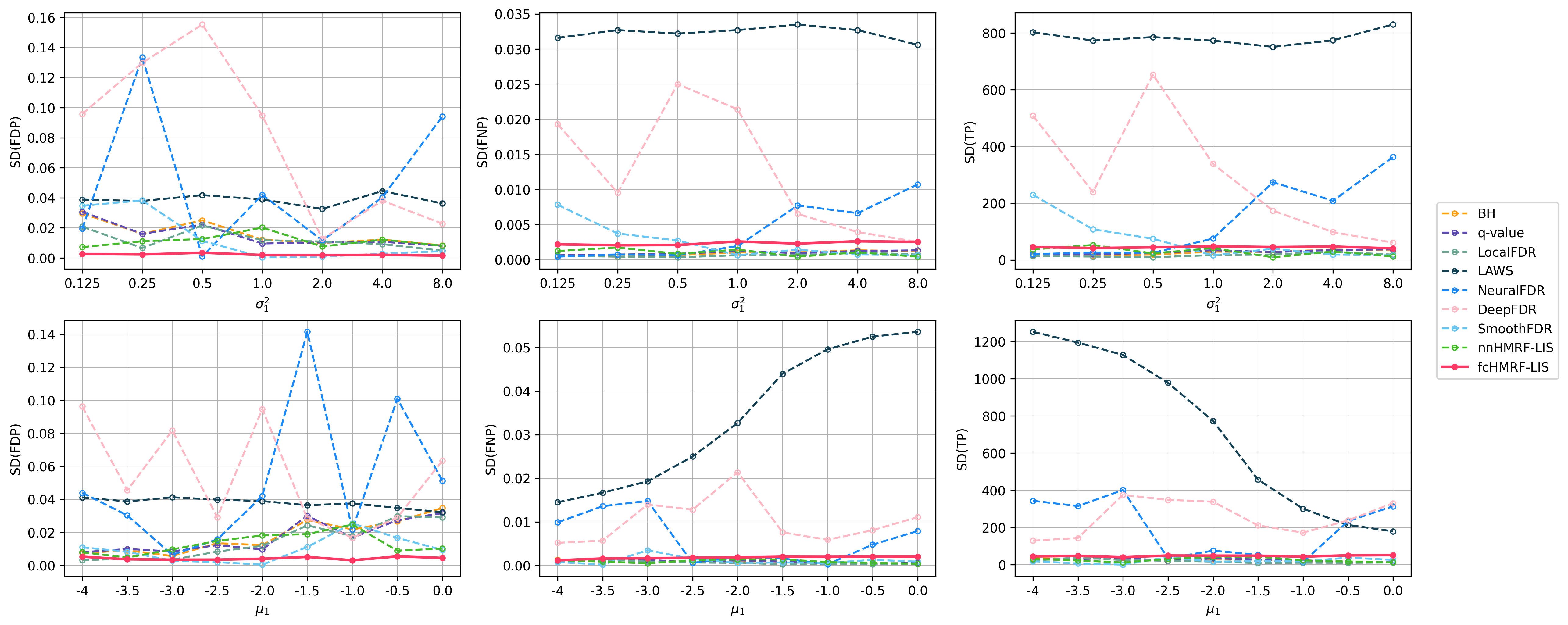}
    \caption{The SDs of FDP, FNP, and TP for 
    the cube with about $10\%$ signal proportion.}
    \label{fig: sim 0.1 p=0.1 std}
  \end{subfigure}

\vskip\baselineskip

  \begin{subfigure}[t]{0.49\textwidth}
    \includegraphics[width=\textwidth]{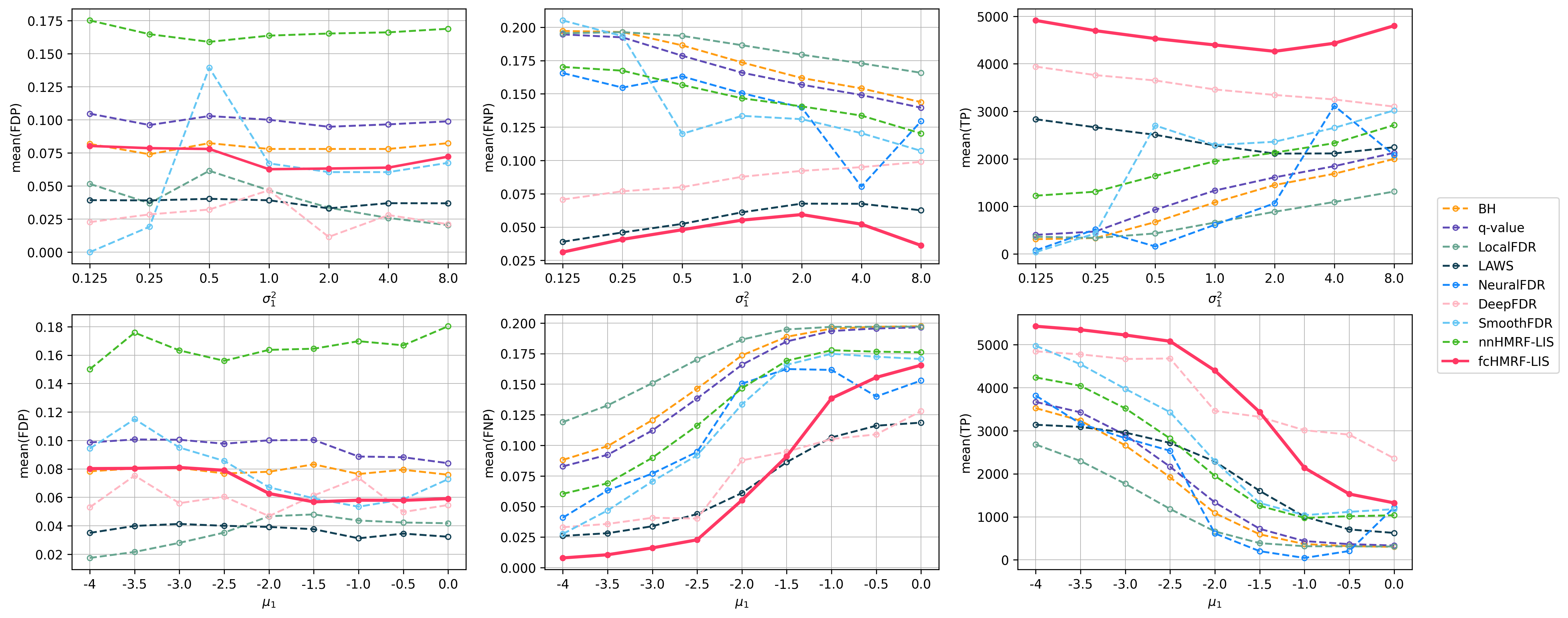}
    \caption{The means of FDP, FNP, and TP for 
    the cube with about $20\%$ signal proportion.
 The mean FDP values of NeuralFDR are omitted because they exceed 0.42.}
    \label{fig: sim 0.1 p=0.2 mean}
  \end{subfigure}
  \hfill
  \begin{subfigure}[t]{0.49\textwidth}
    \includegraphics[width=\textwidth]{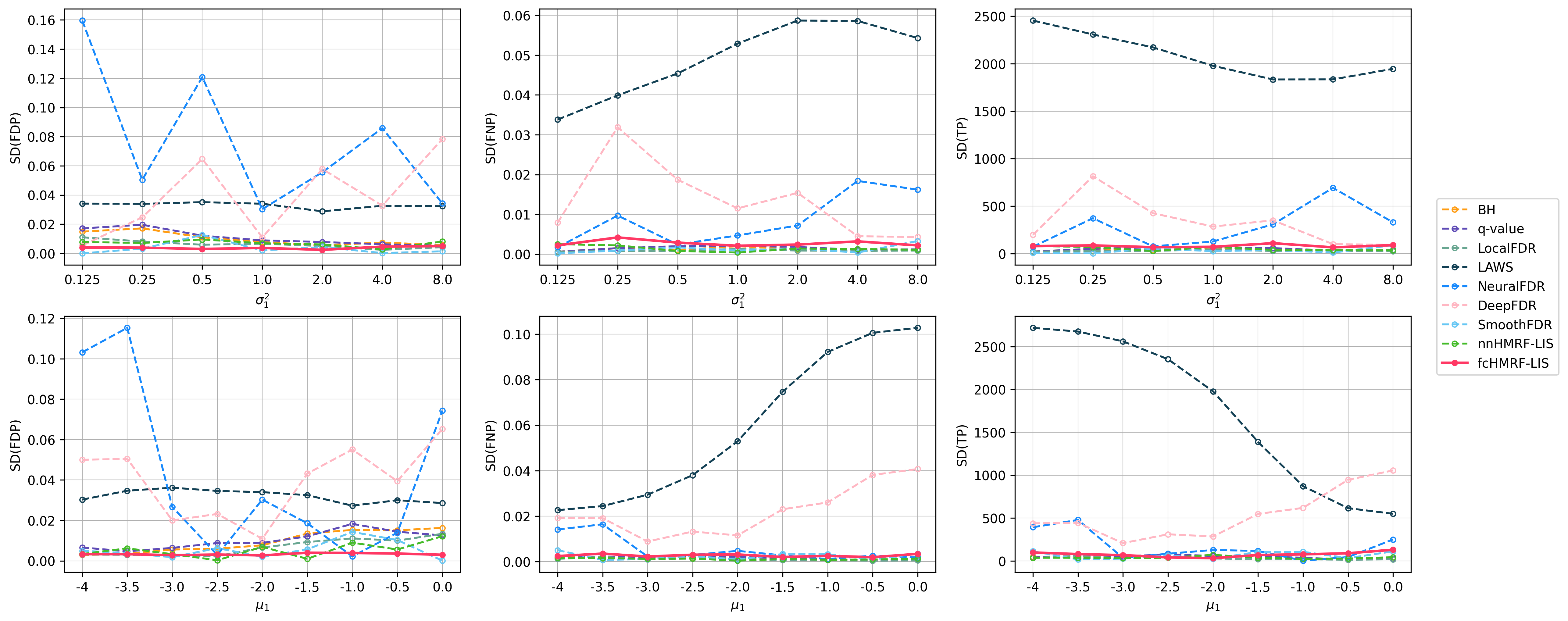}
    \caption{The SDs of FDP, FNP, and TP for 
    the cube with about $20\%$ signal proportion.}
    \label{fig: sim 0.1 p=0.2 std}
  \end{subfigure}

\vskip\baselineskip

  \begin{subfigure}[t]{0.49\textwidth}
    \includegraphics[width=\textwidth]{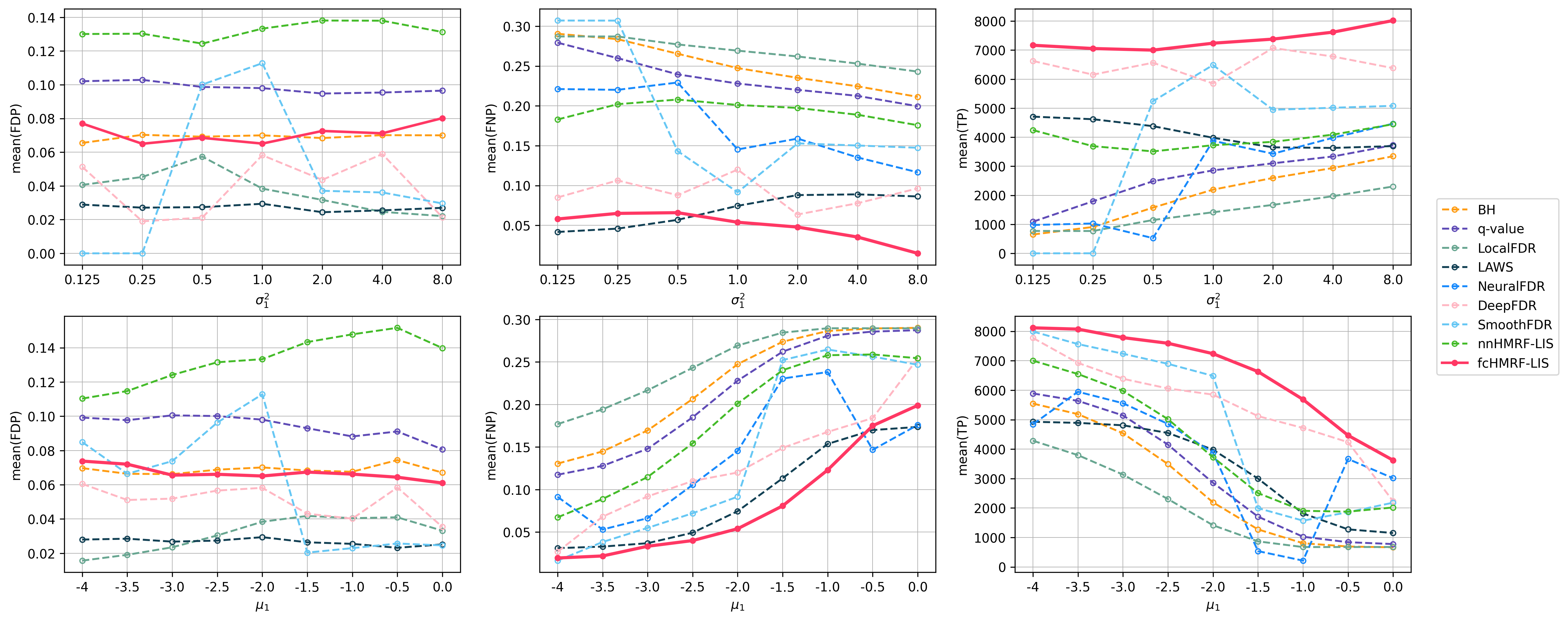}
    \caption{The means of FDP, FNP, and TP for 
    the cube with about $30\%$ signal proportion.
 The mean FDP values of NeuralFDR are omitted because they exceed 0.44.}
    \label{fig: sim 0.1 p=0.3 mean}
  \end{subfigure}
  \hfill
  \begin{subfigure}[t]{0.49\textwidth}
    \includegraphics[width=\textwidth]{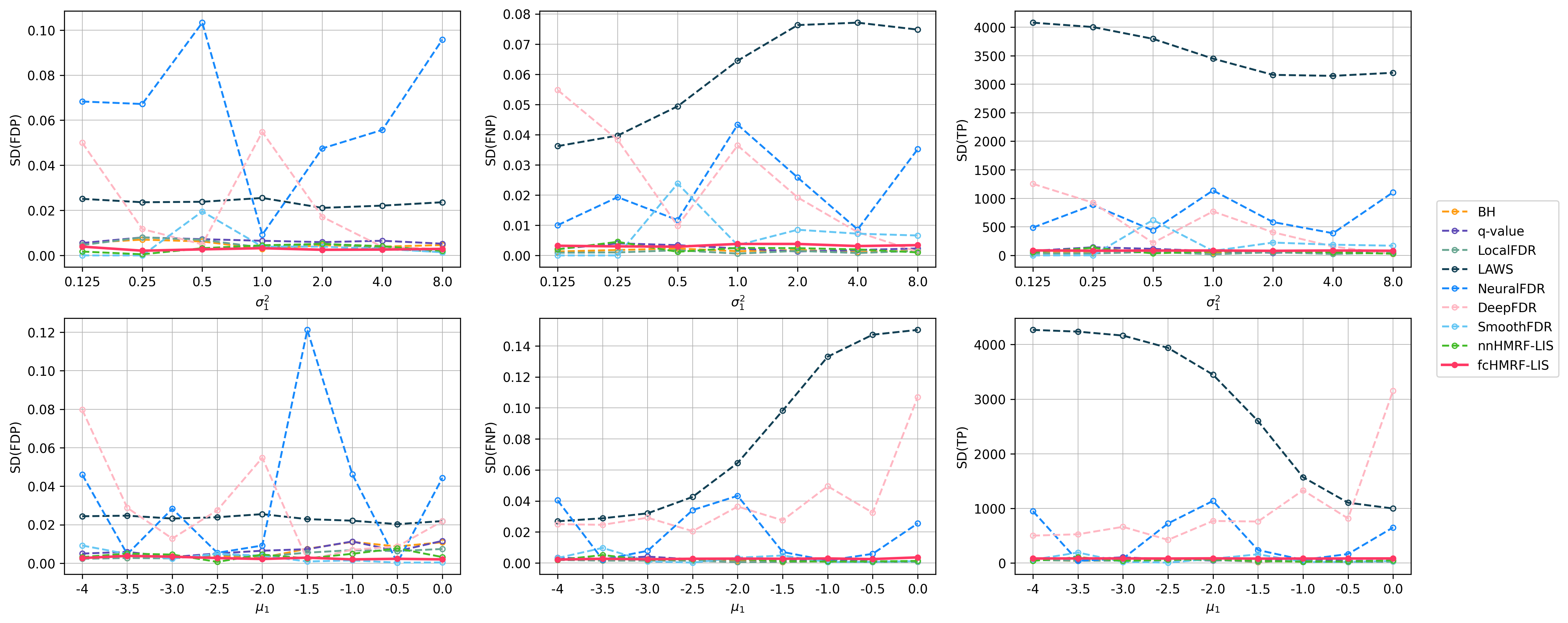}
    \caption{The SDs of FDP, FNP, and TP for 
    the cube with about $30\%$ signal proportion.}
    \label{fig: sim 0.1 p=0.3 std}
  \end{subfigure}
  
    \caption{Simulation results over 50 replications at a nominal FDR level of 0.1.}
    \label{fig: sim 0.1}
\end{figure*}

\subsection{Simulation Settings}\label{sim: setting}

We generate each simulated dataset on a $30\times 30\times 30$ lattice cube,
yielding $m=27{,}000$ voxels.
Since it is computationally prohibitive to generate the true hypothesis states  $\bd{h}=(h_1,\dots, h_m)$
from the fully connected Markov random field in \eqref{eqn: p(h)} using the Gibbs sampler, which requires  $O(m^2B)$ time for $B$ burn-in iterations,
we follow \citep{kim2024deepfdr} and derive $\bd{h}$ from  
the ADNI FDG-PET dataset described in Section~\ref{sec: ADNI dataset}.
Specifically, we use the result of the q-value method at a nominal FDR level of 0.01 for the comparison between the cognitively normal group and 
the group of patients with late mild cognitive impairment  who later converted to AD.
From the brain volume of the q-value result, we randomly crop
three $30\times 30\times 30$ lattice cubes, each containing
approximately 10\%, 20\%, or 30\% of voxels identified as significant. 
In each cube, we assign $h_i= 1$ 
to significant voxels  
and $h_i = 0$ otherwise.
To enhance spatial coherence, isolated signals are reset to $h_i = 0$,
leaving
the signal proportion within each cube nearly unchanged.
The test statistics
$\bd{x}=(x_1,\dots,x_m)$ are generated from the Gaussian mixture model: $x_i|h_i\sim (1-h_i)\mathcal{N}(0,1)+h_i\{0.5\mathcal{N}(\mu_1,\sigma_1^2)+0.5\mathcal{N}(2,1)\}$. We consider 
two types of settings:
(i) varying $\mu_1$ from $-4$ to 0 with  \(\sigma_1^2=1\), 
and (ii) varying $\sigma_1^2$ from 0.125 to 8 with \(\mu_1=-2\). 
The mean difference
$\Delta\mu_i$ used for 
the appearance kernel in
\eqref{eqn: wij}
is estimated from 
the ADNI FDG-PET data
as the
sample mean difference 
adjusted for covariates via
linear regression (see Section~\ref{sec: ADNI dataset}).
In total, we consider
45 simulation settings (15 per signal proportion level). 
 We conduct the nine FDR control methods at a nominal 
 FDR level $\alpha\in \{0.05,0.1\}$
 for 50 independent replications of each simulation setting.

 \subsection{Evaluation Metrics}
 We evaluate the performance of
 the nine FDR control methods in multiple testing by
reporting the means and SDs of the FDP, FNP, and TP over the 50 simulation replications,
where 
the mean FDP and mean FNP
are empirical estimates of the FDR and FNR, respectively.
Since valid FDR control is a fundamental requirement  for FDR control methods, 
ensuring that the mean FDP over replications (i.e., the empirical FDR) is well controlled serves as a prerequisite  for considering the other metrics.
In addition, 
we assess the timing performance 
of each method by reporting the means and SDs of its runtime.

\subsection{Multiple Testing Results}

The multiple testing results are summarized in Figures~\ref{fig: sim 0.05} and~\ref{fig: sim 0.1}.
As shown in subfigures~(a), (c) and (e) of both figures, our proposed fcHMRF-LIS successfully controls the empirical FDR (i.e., the mean FDP) at the nominal level in all simulation settings.
LAWS and LocalFDR also achieve valid empirical FDR control in all settings, and
BH and q-value do so in nearly all settings.
In contrast, 
NeuralFDR and nnHMRF-LIS fail to control the empirical FDR in all settings,
SmoothFDR fails in multiple settings, and DeepFDR fails in several. 
Moreover, fcHMRF-LIS achieves the highest mean TP in 42 out of 45 settings 
at both nominal FDR levels (0.05 and 0.1),
among methods that 
successfully control the empirical FDR 
 in each setting.
The three exceptions occur in  the cube with approximately 20\% signal proportion 
and variance $\sigma_1^2=1$:
for nominal FDR level 0.05 at
 $\mu_1\in \{-1.5, -1,-0.5\}$  (Figure~\ref{fig: sim 0.05 p=0.2 mean}), and for level 0.1 at $\mu_1\in \{-1,-0.5,0\}$  (Figure~\ref{fig: sim 0.1 p=0.2 mean}). 
 In these settings,
 fcHMRF-LIS is
 outperformed by DeepFDR but still ranks second in  mean TP
 among the methods with 
 empirical FDR under control.
In terms of 
empirical FNR
(i.e., the mean FNP),
and still comparing only among methods that successfully control the empirical FDR per setting,
fcHMRF-LIS ranks third (only behind DeepFDR and LAWS) in the  three exceptions at each nominal FDR level,
ranks first in 11 settings at level 0.05 and in 36 settings at level 0.1,
and ranks second in the remaining settings.
Theoretically, if the data exactly follow our fcHMRF model, the LIS-based testing procedure \eqref{def: LIS procedure}
asymptotically achieves the smallest FNR among all valid FDR control methods~\citep{Shu15}. 
Since the simulated data are based on the real ADNI data, they may not perfectly satisfy the fcHMRF model, which may lead to suboptimal performance of fcHMRF-LIS in some settings.

However, a key strength of fcHMRF-LIS 
lies in its stability across replications. 
 In contrast, DeepFDR and LAWS exhibit substantial variability, particularly in the aforementioned settings where fcHMRF-LIS shows suboptimality in mean TP and mean FNP. 
 As shown in subfigures~(b), (d) and~(f) of Figures~\ref{fig: sim 0.05} and~\ref{fig: sim 0.1}, all four recent spatial methods (DeepFDR, LAWS, SmoothFDR, and NeuralFDR) display high variability in FDP, FNP, and/or TP across all or some settings.
%. All FDRs of NeuralFDR are too large, exceeding 0.3. SmoothFDR exhibit good performance for majority of cases, but generally displays objectively higher variations. 
In comparison, fcHMRF-LIS consistently achieves low SDs of 
FDP, FNP, and TP in all settings. 
 The nearest-neighbor-based method,
nnHMRF-LIS, also
 demonstrates strong stability
 in FNP and TP, 
though it fails to  control the empirical FDR. The classical non-spatial methods BH, q-value, and LocalFDR prioritize FDR control but have substantially lower statistical power compared to the spatial methods   fcHMRF-LIS, DeepFDR, and LAWS. Nevertheless, these three classical methods exhibit strong stability, with low variability in FNP and TP across all settings and in FDP in most settings, comparable to that of our fcHMRF-LIS.

\subsection{Timing Performance}\label{sec: simul timing}
We evaluate the timing performance of the nine FDR control methods.
BH, q-value, LocalFDR, SmoothFDR, LAWS, and fcHMRF-LIS are run on a server with 20 Intel Xeon Platinum 8268 CPU cores (2.90GHz, total 64GB RAM). DeepFDR, NeuralFDR, and nnHMRF-LIS 
are executed on the same CPU configuration and an NVIDIA RTX8000 GPU (48GB VRAM). Runtime is measured under the simulation setting with $(\mu_1,\sigma_1^2)=(-2,1)$ and  approximately $ 20\%$ signal proportion at the nominal FDR level of 0.1. 
Table~\ref{table:run_time} reports the mean and SD of  runtime over the 50 simulation replications for each method. 
As expected,
 BH, q-value, and LocalFDR, which are designed for independent tests rather than spatial data, 
achieve the fastest performance, 
each completing in less than 0.14 seconds on average with minimal variability.
Our fcHMRF-LIS ranks seventh in speed, 
with a  mean runtime of 221.14 seconds and an SD of 21.83 seconds.
This is approximately 5/8 the runtime of LAWS and about 1/27 that of NeuralFDR.
In comparison, fcHMRF-LIS takes about 2.4 times the runtime of the simpler nnHMRF-LIS method and about 1.7 times that of SmoothFDR.
However, as the number of tests increases from 27,000 in the simulation setting to 439,758 in the real ADNI dataset, fcHMRF-LIS exhibits significantly better scalability
than nnHMRF-LIS and SmoothFDR,
requiring 
only about 1/5 of the runtime of nnHMRF-LIS
and about 1/14 that of SmoothFDR; see Section~\ref{sec: ADNI time and memory} for further details.

\begin{table}[b!]
\centering
\caption{Mean (SD) of runtime in seconds.}
\label{table:run_time}
\begin{tabular}{l|c|c}
\hline
Method & Simulation & ADNI \\
\hline
BH & 0.050 (0.001) & 0.080 (0.015) \\
q-value & 0.135 (0.003) & 2.203 (0.043) \\
LocalFDR & 0.036 (0.005) & 0.260 (0.036) \\
nnHMRF-LIS & 92.19 (10.23) & 20245 (1987) \\
SmoothFDR & 133.22 (8.16) & 58745 (7397) \\
NeuralFDR & 5995 (442.4) & 92589 (13361) \\
LAWS & 349.38 (29.19) & 538146 (15102) \\
DeepFDR & 7.708 (1.617) & 89.98 (5.17) \\
fcHMRF-LIS & 221.14 (21.83) & 4341 (502.4) \\
\hline
\end{tabular}
\end{table}

\begin{figure*}[p!] % Use [t] or [b] to control vertical placement
\centering
\includegraphics[width=\textwidth]{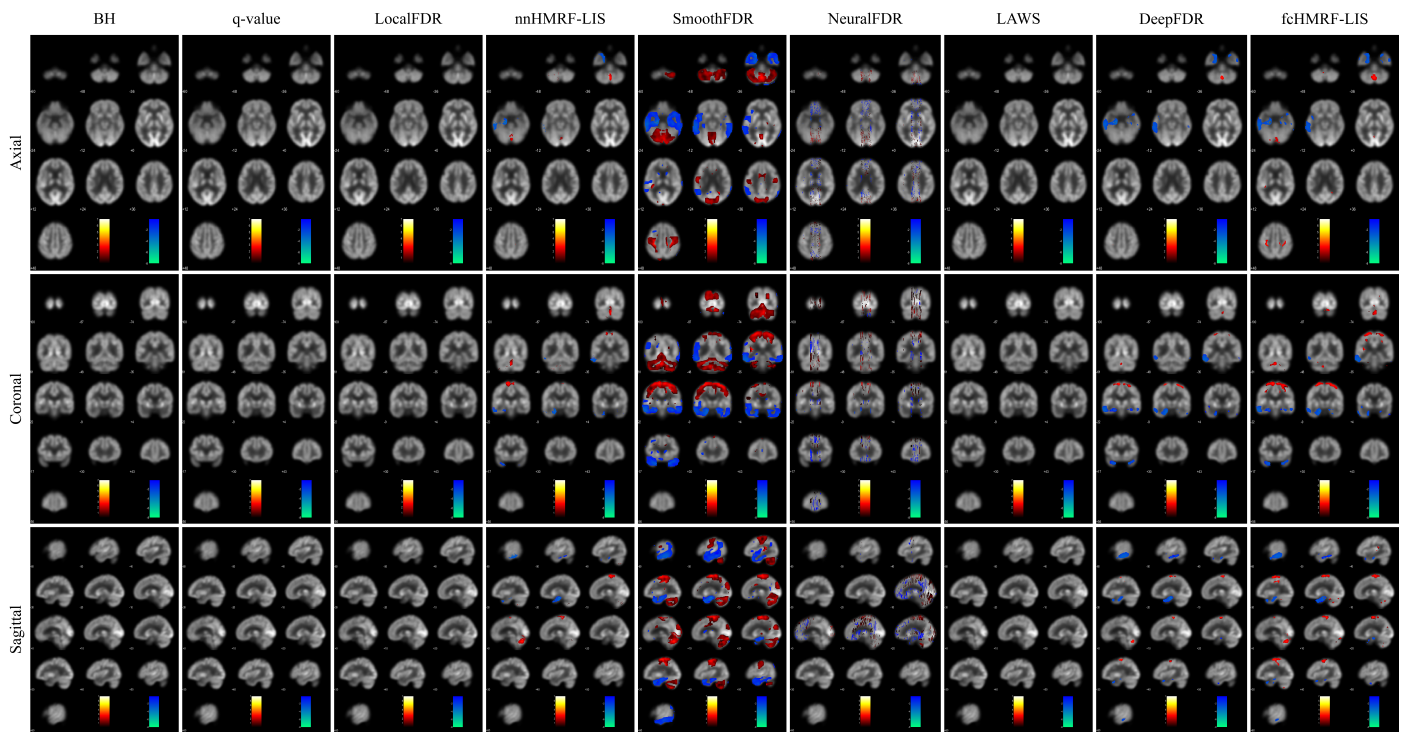}
\caption{The z-statistics of  discoveries for EMCI2AD vs. CN.}
\label{fig: zvalue EMCI2AD}
\end{figure*}

\begin{figure*}[p!] % Use [t] or [b] to control vertical placement
\centering
\includegraphics[width=\textwidth]{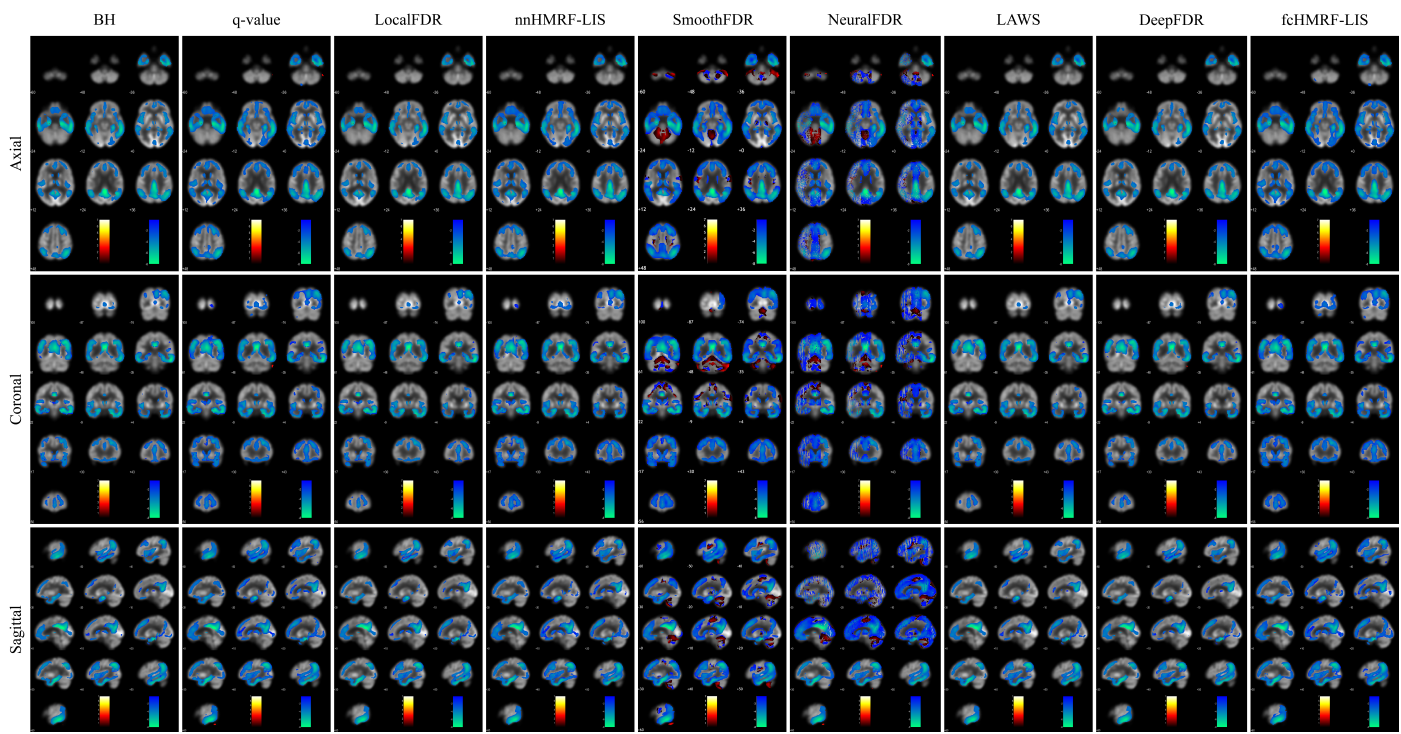}
\caption{The z-statistics of discoveries for LMCI2AD vs. CN.}
\label{fig: zvalue LMCI2AD}
\end{figure*}

\begin{figure*}[h!] % Use [t] or [b] to control vertical placement
\centering
\includegraphics[width=\textwidth]{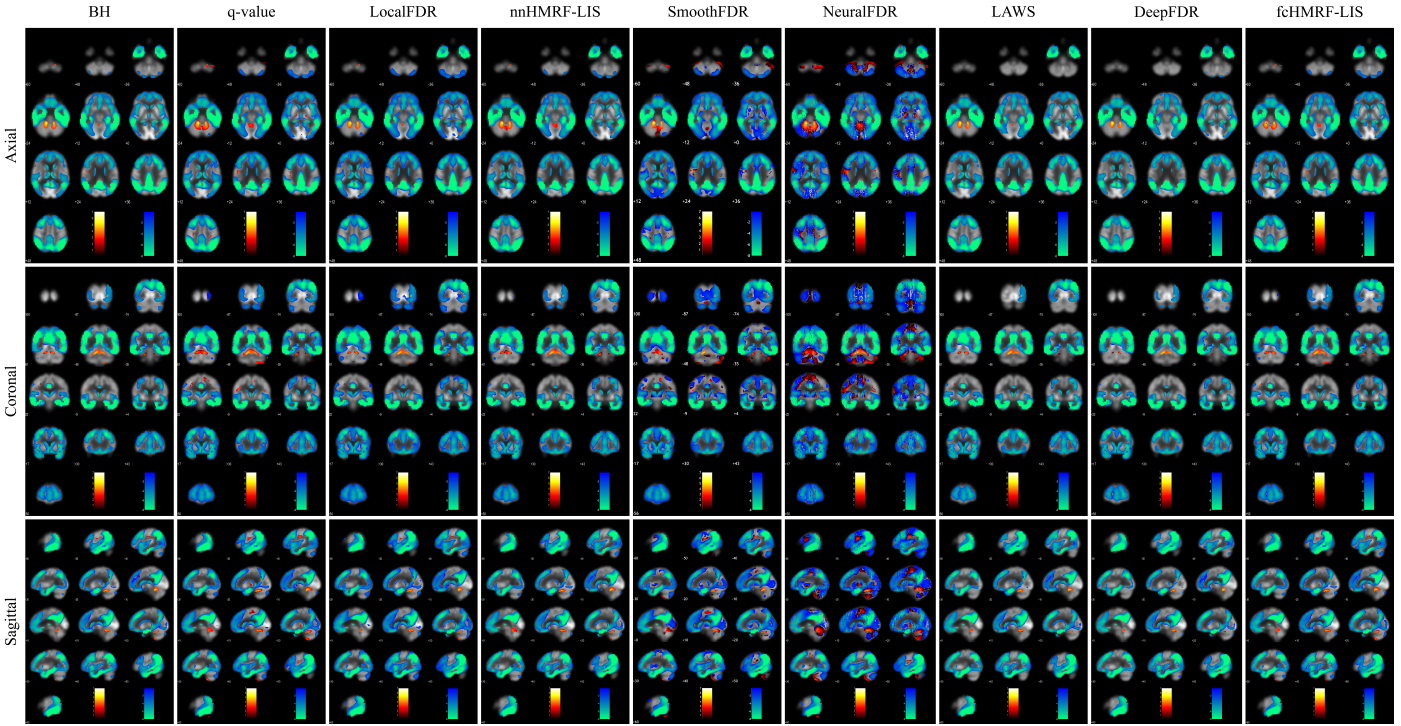}
\caption{The z-statistics of  discoveries for AD vs. CN.}
\label{fig: zvalue AD}
\end{figure*}

\begin{figure*}[h!]
    \centering
    \begin{subfigure}{\textwidth}
        \centering
        \includegraphics[width=\textwidth]{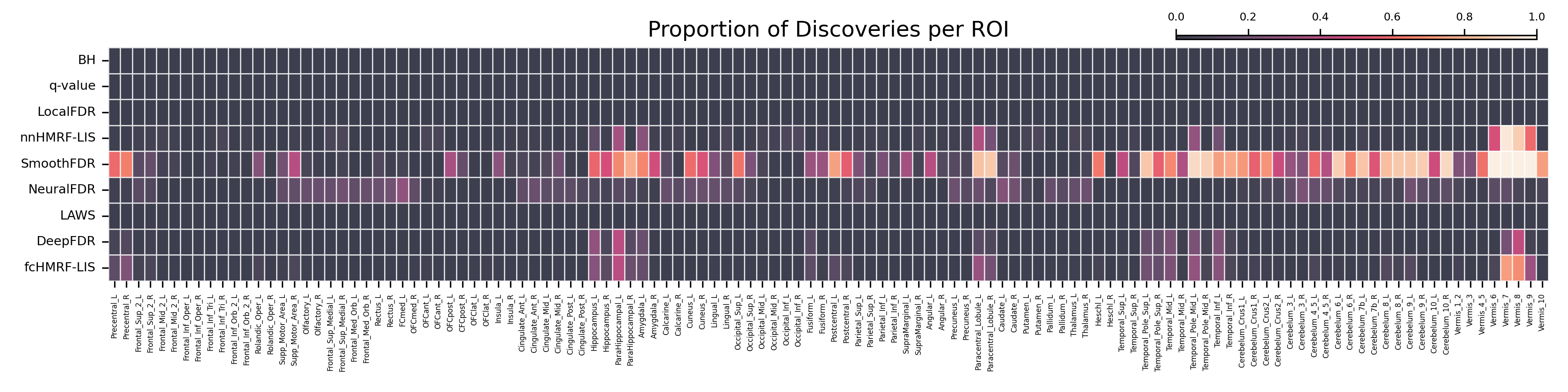}
        \caption{EMCI2AD vs. CN}
        \label{fig:HP EMCI}
    \end{subfigure}
    \begin{subfigure}{\textwidth}
        \centering
        \includegraphics[width=\textwidth]{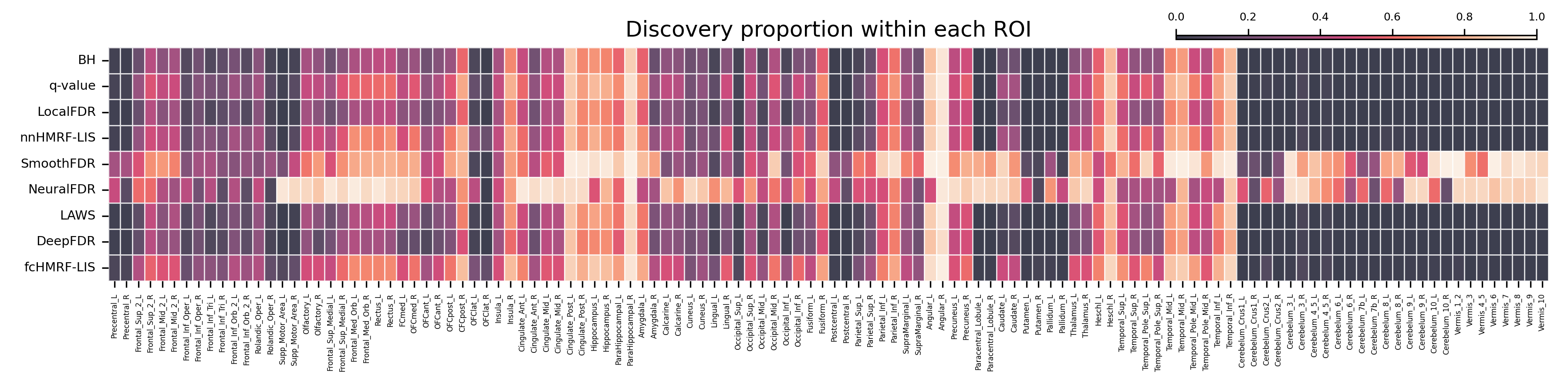}
        \caption{LMCI2AD vs. CN}
        \label{fig:HP LMCI}
    \end{subfigure}
    \begin{subfigure}{\textwidth}
        \centering
        \includegraphics[width=\textwidth]{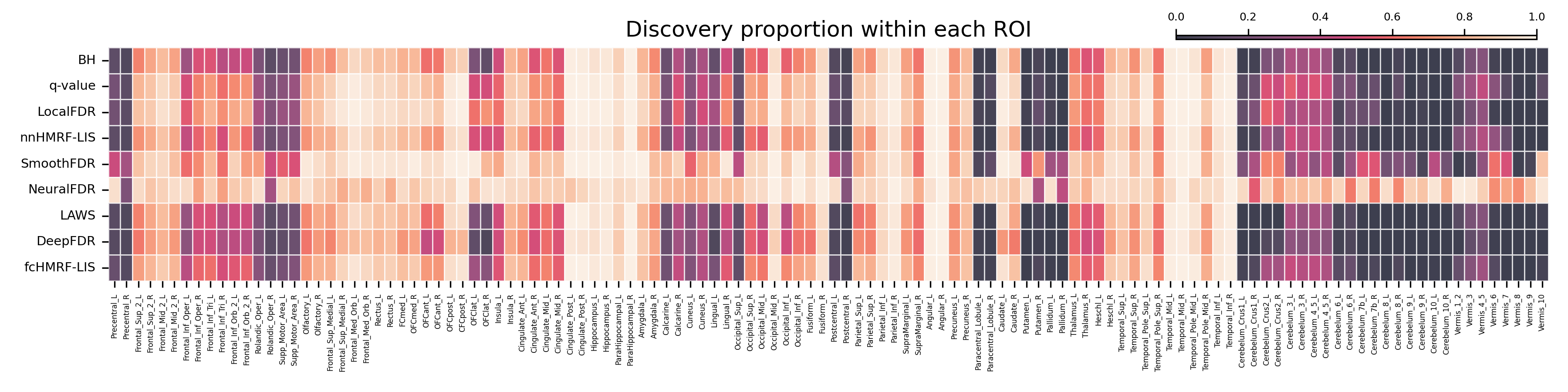}
        \caption{AD vs. CN}
        \label{fig:HP AD}
    \end{subfigure}
    \caption{Heatmaps illustrating the discovery proportion within each ROI 
    for the three comparisons in the ADNI data analysis.}
    \label{fig:heatmap_all}
\end{figure*}

\begin{figure*}[h!] % Use [t] or [b] to control vertical placement
\centering
\includegraphics[width=\textwidth]{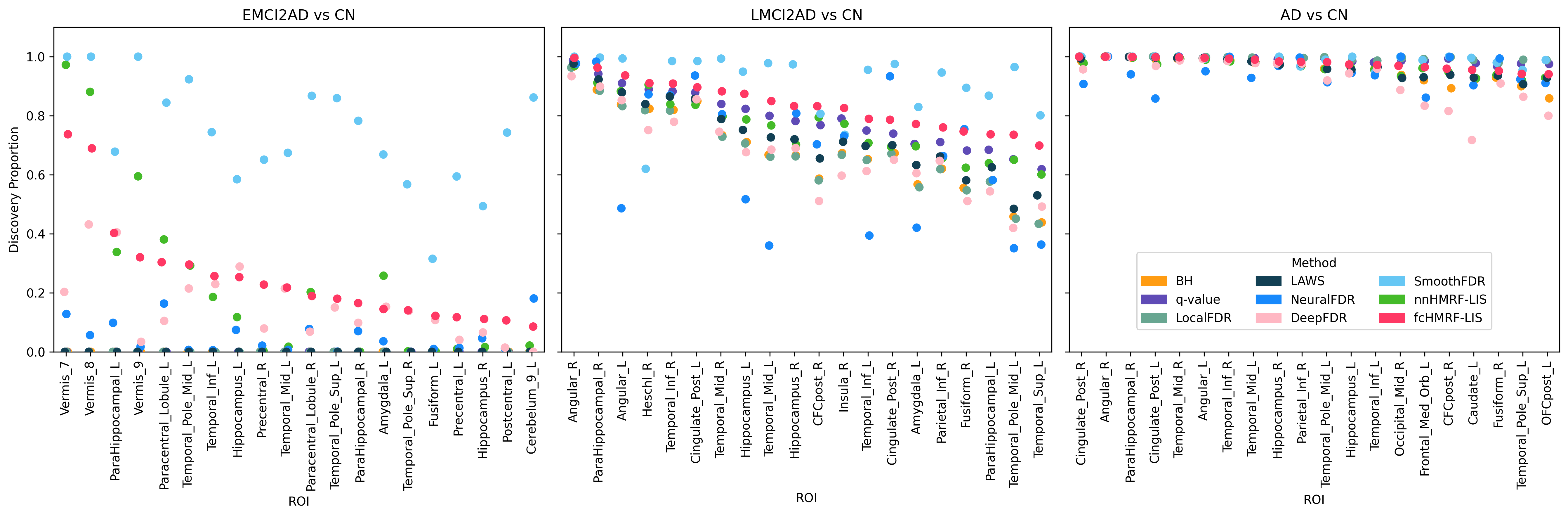}
%\vspace{-0.7cm}
\caption{Top 20 ROIs identified by fcHMRF-LIS and discovery proportions from nine methods for each of the three comparisons in the ADNI data analysis.}
\label{fig: violin}
\end{figure*}

\section{ADNI FDG-PET Data Analysis}\label{sec: real data}

\subsection{Study Aim}
AD is the most common form of dementia, affecting more than 30 million people worldwide, and there is still no curative treatment~\citep{WHO2025}.
FDG-PET, which measures brain glucose metabolism, is  widely used to support the clinical diagnosis of AD
and to monitor its progression~\citep{ou2019fdg}. 
Because glucose metabolism in the brain typically declines with AD, differences across disease stages can be examined by testing voxel-wise mean differences in SUVR. This yields a high-dimensional spatial multiple-testing problem.
In this study, we apply 
voxel-wise multiple testing
to 
compare the mean SUVR difference between the cognitively normal (CN) group and each of the following three groups: early mild cognitive impairment (MCI) patients with conversion to AD (EMCI2AD), late MCI patients with conversion to AD (LMCI2AD), and the AD group.
We evaluate and compare the nine FDR control  methods given in Section~\ref{sec: method compare}.

\subsection{ADNI FDG-PET Data}\label{sec: ADNI dataset} 
We use the FDG-PET brain image dataset 
obtained from the ADNI database (\url{https://adni.loni.usc.edu}).
The ADNI was launched in
2003 as a public-private partnership, led by Principal Investigator Michael W. Weiner,
MD. The primary goal of ADNI has been to test whether serial magnetic resonance imaging (MRI), positron emission tomography (PET), other biological markers, and clinical and
neuropsychological assessment can be combined to measure the progression of MCI and early AD. 
We use  baseline
3D FDG-PET scans from 742 subjects,
including 
286 CN subjects, 42 EMCI2AD patients, 175 LMCI2AD patients, and 239 AD patients. All scans 
are preprocessed using 
the Clinica software \citep{routier2021clinica},
which includes spatial normalization to the MNI IXI549Space template and intensity normalization based on average uptake in the pons region.
We consider the 120
regions of interest (ROIs) defined by the AAL2 atlas~\citep{rolls2015implementation}, resulting in a total of $m=439{,}758$ voxels. At each voxel, we fit a linear regression model with SUVR as the response,  and dummy variables for the EMCI2AD, LMCI2AD, and AD groups as predictors, using the CN group as the reference. 
The model adjusts
for age, gender, race, ethnicity, education, marital status, and APOE4 status by including them as covariates. 
From this model, we obtain 
 the voxel-level t-statistics and p-values for 
 the regression coefficients $\{\hat{\beta}_{\text{EMCI2AD},i},\hat{\beta}_{\text{LMCI2AD},i},\hat{\beta}_{\text{AD},i}\}_{i=1}^m$,
corresponding to the dummy variables for the three AD-related stage groups.
These coefficients represent 
the three comparisons: EMCI2AD vs. CN, LMCI2AD vs. CN, and AD vs. CN.
The voxel-level mean difference 
between an AD-related stage and CN is estimated by
$\Delta\hat{\mu}_i = \hat{\beta}_{\text{stage},i}$.
All t-statistics 
are converted to z-statistics for use in FDR control methods that require them as input.

\subsection{Multiple Testing Results}
\label{sec:realdata}
Nominal FDR levels of $\alpha = 0.05$, 0.01, and  0.005 are applied to the EMCI2AD vs. CN, LMCI2AD vs. CN, and AD vs. CN comparisons, respectively. 
Stricter nominal levels are used for comparisons involving later disease stages to better control the number of false discoveries.
Table~\ref{tab:discovery_stats_supp} summarizes the number of discoveries made by each method and the proportions of discoveries with p-values exceeding the nominal level and 0.05.
Figures~\ref{fig: zvalue EMCI2AD}--\ref{fig: zvalue AD} present  the z-statistics of discoveries  for each method. Most detected brain areas exhibit hypometabolism, with affected areas expanding as AD progresses. 
This trend is further illustrated in Figure~\ref{fig:heatmap_all}, which shows heatmaps of the discovery proportion  in each ROI for all methods.

The EMCI2AD vs. CN comparison is the most challenging among the three comparisons, but is also the most critical for enabling early detection and timely therapeutic intervention in AD. In this comparison, BH, q-value, LocalFDR, and LAWS yield no discoveries. In fact, the three classical non-spatial methods BH, q-value, and LocalFDR fail to produce any discoveries until the nominal FDR level is raised to 0.8. As shown in Table~\ref{tab:discovery_stats_supp}, nnHMRF-LIS, SmoothFDR, NeuralFDR, DeepFDR, and fcHMRF-LIS identify 13587, 141656, 25675, 11341, and 20758 discoveries,  respectively. However, 
a substantial proportion of discoveries made by nnHMRF-LIS (55.86\%), SmoothFDR (80.48\%), and NeuralFDR (95.10\%) have 
p-values greater than the nominal level of 0.05. While a discovery with a p-value above the nominal level is not  necessarily a false signal, a large proportion of such discoveries can raise concerns about the validity of these methods in controlling the FDR. In contrast, 
DeepFDR and fcHMRF-LIS report only 
 2.14\% and 4.75\% of discoveries
 with p-values greater than 0.05, respectively,
 suggesting more credible results. 
Among all methods except SmoothFDR and NeuralFDR, fcHMRF-LIS  achieves the largest number of discoveries.

In the LMCI2AD vs. CN comparison, fcHMRF-LIS identifies  181209 discoveries, with
 13.38\% having p-values greater than the nominal level of 0.01 and 0.82\% exceeding 0.05.
Again, it produces the most discoveries aside from SmoothFDR and NeuralFDR, both of which yield more than 42\% of discoveries with
 p-values greater than 0.01 and over 27\% exceeding 0.05.
The distribution of discoveries from fcHMRF-LIS  is similar to those from the six other methods excluding SmoothFDR and NeuralFDR, 
as shown in Figures~\ref{fig: zvalue LMCI2AD}
and~\ref{fig:HP LMCI}. 

In the AD vs. CN comparison,  fcHMRF-LIS detects 260229 discoveries, with 3.36\% having p-values greater than
the nominal level of 0.005 and 0.04\% exceeding  0.05. Its distribution of discoveries is again similar to those obtained by the six methods other than SmoothFDR and NeuralFDR,
as shown in Figures~\ref{fig: zvalue AD}
and~\ref{fig:HP AD}. In contrast, 
SmoothFDR and NeuralFDR report
24.40\% and 37.54\% of discoveries
with p-values greater than 0.005, and 13.35\% and 25.21\% 
exceeding 0.05, respectively.

 Across all three comparisons, SmoothFDR and NeuralFDR appear to overestimate the signals, as evidenced by the large proportions of discoveries with 
p-values exceeding the nominal levels and even 0.05. 
Additionally, Figures~\ref{fig: zvalue EMCI2AD}–\ref{fig: zvalue AD} reveal that SmoothFDR tends to produce oversmoothed results, while NeuralFDR yields discoveries with reduced spatial coherence.

\iffalse

Vermis_7,8,9 (7,9: The cerebellum in Alzheimer’s disease: evaluating its role in cognitive decline
8: Whole-Brain Structure-Function Coupling Abnormalities in Mild Cognitive Impairment: A Study Combining ALFF and VBM)

ParaHippocampal_L, R        x
Paracentral_Lobule_L, R   （Altered Cerebro-Cerebellar Limbic Network in AD Spectrum: A Resting-State fMRI Study）
Temporal_Pole_Mid_L   &  Temporal_Pole_Sup_L, R   (%Temporal_Pole_Sup_L: Amnestic Mild Cognitive Impairment Is Associated With Frequency-Specific Brain Network Alterations in Temporal Poles

Revealing heterogeneity in mild cognitive impairment based on individualized structural covariance network Better
)
Temporal_Inf_L         x     
Hippocampus_L, R      x
Precentral_R,L    （L: Functional and structural alterations of dorsal attention network in preclinical and early-stage Alzheimer's disease

Abnormal Structural Brain Connectome in Individuals with Preclinical Alzheimer’s Disease. Better）

Temporal_Mid_L     x
Amygdala_L  x  (Amidst an amygdala renaissance in Alzheimer’s disease )
Fusiform_L        x   
Postcentral_L   (Changes in Cortical Activation during Retrieval of Clock Time Representations in Patients with Mild Cognitive Impairment and Early Alzheimer’s Disease)
Cerebelum_9_L (Differences Changes in Cerebellar Functional Connectivity Between Mild Cognitive Impairment and Alzheimer's Disease: A Seed-Based Approach

The cerebellum in Alzheimer’s disease: evaluating its role in cognitive decline
)

\fi

Figure~\ref{fig: violin}
shows the top 20 ROIs for each of the three comparisons, ranked by discovery proportions from fcHMRF-LIS, along with corresponding values from all nine methods.
For the EMCI2AD vs. CN comparison, discovery proportions vary widely across the nine methods. This variability  decreases substantially for the LMCI2AD vs. CN comparison and becomes even smaller for the AD vs. CN comparison, indicating increasing agreement among methods on the most affected ROIs in later disease stages.
In the most challenging 
EMCI2AD vs. CN comparison, 
as noted earlier, our fcHMRF-LIS
yields the  largest number of high-quality discoveries.
Its top 20 ROIs include
lobules VII, VIII, and IX of the vermis, 
bilateral parahippocampal gyri,
bilateral hippocampi,
bilateral paracentral lobules, 
bilateral precentral gyri, 
bilateral superior temporal poles,
left middle temporal pole, 
left middle temporal gyrus,
left inferior temporal gyrus, 
left amygdala,
left fusiform gyrus, 
left postcentral gyrus, and 
left lobule IX of the cerebellar hemisphere.
These 20 ROIs are not detected in the FDG-PET data by any of the three classical methods (BH, q-value, and LocalFDR) or by the recent spatial method LAWS.
However, 
prior studies using structural or functional 
 MRI data have shown that 
 these regions are affected in  early stages of AD~\citep{jacobs2018cerebellum,zhao2023whole,echavarri2011atrophy,apostolova2010subregional,qi2019altered,pereira2018abnormal,wei2025revealing,convit2000atrophy,stouffer2024amidst,leyhe2009changes,tang2021differences},
providing strong support for the neurobiological relevance and enhanced sensitivity of the fcHMRF-LIS method.

\iffalse

Among these top 20 ROIs, for all three comparisons, fcHMRF-LIS discovers signals in the right parahippocampal gyrus, left and right hippocampus, left mid temporal gyrus, and left inferior temporal gyrus. Additionally, left and right angular gyrus, left and right posterior cingulate cortex, and right fusiform gyrus consistently appear across LMCI2AD and AD vs. CN comparisons, with over 50\% discovery proportions averaged across all methods. These ROIs have been linked to brain atrophy due to AD \citep{echavarri2011atrophy, THOMPSON20041754, convit2000atrophy, galton2001differing, raffaella_2015, HUNT2007147}. 

The notable difficulty of the EMCI2AD vs. CN task compared to LMCI2AD vs. CN or AD vs. CN is evident. The variation in the discovery proportions found by the nine methods for the top 20 ROIs is large for EMCI2AD vs. CN comparison, but noticeably decreases for LMCI2AD vs. CN, and even more for AD vs. CN, indicating greater coherence and confidence in the detected differences. This suggests that the complex dynamics and heterogeneity of functional patterns during the EMCI phase make it challenging to confidently identify EMCI-associated ROIs utilizing FDG PET data only. These findings underscore the need for continued advancement in spatial multiple hypothesis testing through multimodal integration, longitudinal analyses, and more sensitive statistical modeling to enhance our understanding of early-stage AD.

\fi

\subsection{Time and Memory Efficiency}\label{sec: ADNI time and memory}

We execute the nine FDR control methods on the same computational resource as specified in Section~\ref{sec: simul timing}. 
Table~\ref{table:run_time}  reports 
each method's
 mean and SD of runtime over the three comparisons in the ADNI data analysis. 
Our fcHMRF-LIS 
achieves  a mean runtime of 4341 seconds
(SD = 502.4 seconds),
ranking fifth overall, only
behind the three non-spatial methods (BH, q-value, and LocalFDR) and the deep learning-based method DeepFDR.
As expected,
the three non-spatial  methods 
are the fastest, each with a mean runtime under 2.3 seconds (SD under 0.05 seconds), as they do not account for spatial dependence among tests.
DeepFDR benefits from GPU acceleration and, like most deep learning models, is expected to exhibit substantially longer runtime when executed without GPU support.
In contrast, our fcHMRF-LIS 
runs entirely on CPUs 
and is significantly more efficient than other spatial methods, requiring only about 
  1/5 the runtime of nnHMRF-LIS,
1/14  of SmoothFDR, 1/21  of NeuralFDR,
and 1/124  of LAWS.

Table~\ref{table:memory_usage} presents the mean and SD of  peak memory usage 
for each method
over the three comparisons.
The non-spatial methods BH, q-value, and LocalFDR exhibit minimal CPU memory usage
(mean $\le$ 0.56 GB, SD $\le$ 0.07 GB)
and require no GPU. 
Between the two
deep learning-based methods,
DeepFDR shows the highest GPU memory usage (mean = 33.75 GB, SD = 0.91 GB) due to its W-Net  architecture, whereas NeuralFDR, based on a multilayer perceptron, consumes less GPU memory (mean = 15.39 GB, SD = 0.81 GB) but requires substantially longer runtime.
 Our fcHMRF-LIS demonstrates moderate peak CPU memory usage (mean = 2.91 GB, SD = 0.59 GB) while operating entirely on CPUs. All methods can be executed on standard hardware, with minimal memory requirements for the non-spatial methods and moderate demands for the spatial methods. Overall, our fcHMRF-LIS achieves excellent computational efficiency and scalability for high-dimensional data analysis.

\begin{table}[ht]
\centering
\caption{Summary of discoveries  for the three comparisons in ADNI data analysis. Abbreviations: Disc. = discoveries; w/ = with; $p$ = p-value.}
\scriptsize
\setlength{\tabcolsep}{4pt}
\renewcommand{\arraystretch}{1.1}
\resizebox{\columnwidth}{!}{%
\begin{tabular}{lccc}
\toprule
Method & $\#\{\text{Disc.}\}$ & $\dfrac{\# \{\text{Disc. w/ } p {>} \alpha\}}{\# \{\text{Disc.}\}}$ & $\dfrac{\# \{\text{Disc. w/ }  p {>} 0.05\}}{\# \{\text{Disc.}\}}$  \\
\midrule
\multicolumn{4}{c}{\underline{EMCI2AD vs. CN ($\alpha = 0.05$; $\#\{p  \le \alpha\}= 27645$)}} \\
BH         &      0 &     --    & -- \\
q-value    &      0 &     --    & -- \\
LocalFDR   &      0 &     --    & -- \\
nnHMRF-LIS &  13587 &  0.5586   & 0.5586 \\
SmoothFDR  & 141656 &  0.8048   & 0.8048 \\
NeuralFDR  &  25675 &  0.9510   & 0.9510 \\
LAWS       &      0 &     --    & -- \\
DeepFDR    &  11341 &  0.0214   & 0.0214 \\
fcHMRF-LIS &  20758 &  0.0475   & 0.0475 \\
\\[-0.8em]
\multicolumn{4}{c}{\underline{LMCI2AD vs. CN ($\alpha = 0.01$; $\#\{ p \le \alpha\}= 157075$)}} \\
BH         & 119374 & 0.0000    & 0.0000 \\
q-value    & 159516 & 0.0153    & 0.0000 \\
LocalFDR   & 117783 & 0.0000    & 0.0000 \\
nnHMRF-LIS & 155592 & 0.0400    & 0.0093 \\
SmoothFDR  & 256271 & 0.4240    & 0.2797 \\
NeuralFDR  & 255743 & 0.5375    & 0.4232 \\
LAWS       & 125562 & 0.0007    & 0.0000 \\
DeepFDR    & 115325 & 0.0034    & 0.0000 \\
fcHMRF-LIS & 181209 & 0.1338    & 0.0082 \\
\\[-0.8em]
\multicolumn{4}{c}{\underline{AD vs. CN ($\alpha = 0.005$; $\#\{p  \le \alpha\}= 251543$)}} \\
BH         & 240571 & 0.0000    & 0.0000 \\
q-value    & 282201 & 0.1086    & 0.0000 \\
LocalFDR   & 290968 & 0.1390    & 0.0000 \\
nnHMRF-LIS & 252794 & 0.0227    & 0.0026  \\
SmoothFDR  & 326667 & 0.2440    & 0.1335 \\
NeuralFDR  & 375010 & 0.3754    & 0.2521  \\
LAWS       & 236787 & 0.0042    & 0.0000  \\
DeepFDR    & 227756 & 0.0063    & 0.0002  \\
fcHMRF-LIS & 260229 & 0.0336    & 0.0004 \\
\bottomrule
\end{tabular}%
}
\label{tab:discovery_stats_supp}
\end{table}

\begin{table}[h!]
\centering
\caption{Mean (SD) of peak memory usage (in GB) for  the ADNI data analysis.}
\label{table:memory_usage}
\begin{tabular}{l|c|c}
\hline
Method & RAM (CPU) & VRAM (GPU) \\
\hline
BH & 0.469 (0.052) & -- \\
q-value & 0.471 (0.056) & -- \\
LocalFDR & 0.557 (0.062) & -- \\
nnHMRF-LIS & 0.911 (0.085) & 0.302 (0.022) \\
SmoothFDR & 3.380 (0.401) & -- \\
NeuralFDR & 0.766 (0.068) & 15.385 (0.814) \\
LAWS & 2.413 (0.349) & -- \\
DeepFDR & 3.026 (0.570) & 33.752 (0.908) \\
fcHMRF-LIS & 2.909 (0.592) & -- \\
\hline
\end{tabular}
\end{table}

\section{Conclusion}
\label{sec:conclusion} 
In this paper, we propose fcHMRF-LIS, a novel spatial FDR control method  for large-scale voxel-wise multiple testing in neuroimaging studies. 
The method is designed to
address three major limitations of existing methods: inadequate modeling of spatial dependence, high variability in FDP and FNP, and limited computational 
 scalability. To overcome these challenges,  fcHMRF-LIS integrates 
 the LIS-based testing procedure with our newly designed
 fcHMRF, enabling effective and efficient inference under complex spatial structures.

We develop an efficient EM algorithm for parameter estimation using the mean-field approximation, CRF-RNN technique, and permutohedral lattice filtering, reducing time complexity from quadratic to linear in the number of tests. In extensive simulations, fcHMRF-LIS achieves superior FDR control, lower FNR, higher TP, and reduced variability across replications, compared to eight existing methods, including classical, local smoothing-based, and deep learning-based approaches.

When applied to ADNI FDG-PET data containing 439,758 voxels of interest, fcHMRF-LIS completes each of  three  comparisons in about 1.2 hours on a 20-core CPU server without GPU acceleration.
This is significantly faster than LAWS, which requires about 149.5 hours on the same server, and more efficient than nnHMRF-LIS and NeuralFDR, which take about 5.6 and 25.7 hours, respectively, using a GPU. 
Notably, in the most  challenging EMCI2AD vs. CN comparison, fcHMRF-LIS identifies 20 ROIs that are  supported by prior structural and functional MRI studies, but missed by classical FDR control methods using the FDG-PET data. This highlights both the sensitivity and neurobiological relevance of fcHMRF-LIS.

These results demonstrate that fcHMRF-LIS is a powerful, stable, and scalable method for voxel-wise multiple testing in neuroimaging. Future work will explore extensions of this framework to other domains with complex spatial or sequential structures, such as GWAS SNP data  and time series data.

\section*{Acknowledgements}
Dr. Shu's research was partially supported by  NYU GPH Goddard Award and NYU GPH Research Support Grant. 
Dr. de~Leon's research was partially supported by the grants R01 AG12101, R01 AG022374, R01 AG13616, RF1 AG057570, and R56 AG058913
from the National Institutes of Health (NIH).
The content is solely the responsibility of the authors and does not necessarily represent the official views of the NIH.

This work was supported in part through the NYU IT High Performance Computing resources, services, and staff \mbox{expertise.}

Data collection and sharing for this project was funded by the Alzheimer's Disease
Neuroimaging Initiative (ADNI) (National Institutes of Health Grant U01 AG024904) and
DOD ADNI (Department of Defense award number W81XWH-12-2-0012). ADNI is funded
by the National Institute on Aging, the National Institute of Biomedical Imaging and
Bioengineering, and through generous contributions from the following: AbbVie, Alzheimer’s
Association; Alzheimer’s Drug Discovery Foundation; Araclon Biotech; BioClinica, Inc.;
Biogen; Bristol-Myers Squibb Company; CereSpir, Inc.; Cogstate; Eisai Inc.; Elan
Pharmaceuticals, Inc.; Eli Lilly and Company; EuroImmun; F. Hoffmann-La Roche Ltd and
its affiliated company Genentech, Inc.; Fujirebio; GE Healthcare; IXICO Ltd.; Janssen
Alzheimer Immunotherapy Research \& Development, LLC.; Johnson \& Johnson
Pharmaceutical Research \& Development LLC.; Lumosity; Lundbeck; Merck \& Co., Inc.;
Meso Scale Diagnostics, LLC.; NeuroRx Research; Neurotrack Technologies; Novartis
Pharmaceuticals Corporation; Pfizer Inc.; Piramal Imaging; Servier; Takeda Pharmaceutical
Company; and Transition Therapeutics. The Canadian Institutes of Health Research is
providing funds to support ADNI clinical sites in Canada. Private sector contributions are
facilitated by the Foundation for the National Institutes of Health (www.fnih.org). The grantee
organization is the Northern California Institute for Research and Education, and the study is
coordinated by the Alzheimer’s Therapeutic Research Institute at the University of Southern
California. ADNI data are disseminated by the Laboratory for Neuro Imaging at the
University of Southern California.

\appendix

\section{Implementation Details of  Methods}
\label{app sec: implement}

This appendix provides implementation details for all methods used in our simulation studies and ADNI data analysis. Python implementations were used whenever available; otherwise, R versions were employed. All numerical studies were conducted using Python~3.12.5 and R~4.2.1.

\begin{itemize}[leftmargin=1em]
    \item \textbf{BH:} Takes a 1D sequence of p-values as input. We used the  Python package \texttt{statsmodels} (v0.12.2), available at 
\url{https://www.statsmodels.org}. 

    \item \textbf{q-value:} Takes a 1D sequence of p-values as input. We used the Python package \texttt{multipy} (v0.16), available at \url{https://github.com/puolival/multipy}. 
    
    \item \textbf{LocalFDR:} Takes a 1D sequence of z-statistics as input. Implemented using the Python package \texttt{statsmodels} (v0.12.2),  available at 
\url{https://www.statsmodels.org}. 

    \item \textbf{nnHMRF-LIS:} Take a 3D volume of z-statistics as input. 
    We used the GPU-enabled Python package  
 from
    \url{https://github.com/shu-hai/nnHMRF-LIS}. 
    In simulations, a single HMRF was used for the  $30 \times 30 \times 30$ lattice cube. 
    In ADNI data analysis, each ROI was modeled with a separate HMRF, and the LIS values were then pooled for testing.
     
    \item \textbf{SmoothFDR:} Take a 3D volume of z-statistics as input. We used the Python package from \url{https://github.com/tansey/smoothfdr}, with 20 sweeps. 
    
    \item \textbf{NeuralFDR:} Takes p-values as input along with 3D spatial coordinates as covariates. 
    Implemented using
    the Python package from \url{https://github.com/fxia22/NeuralFDR}. 
        For simulations, default parameters were used.
    For ADNI data,
we modified the code to enable mini-batch training (batch size=27,000) due to GPU memory constraints,
and 
set the parameters \texttt{n-init=3} and \texttt{num-iterations=200}.
    
    \item \textbf{LAWS:} Takes a 3D volume  of p-values as input. 
    We used the R code provided in the supplementary materials of the LAWS paper~\citep{laws}, available
 at \url{https://doi.org/10.1080/01621459.2020.1859379}. 
 
    \item \textbf{DeepFDR:} Takes a 3D volume of test statistics (z-statistics for simulations and t-statistics for ADNI data) and corresponding 3D volume of p-values as input. We used the Python package from \url{https://github.com/kimtae55/DeepFDR}. 
For simulations, 
the  $30\times 30\times 30$ volume
was zero-padded to  $32\times 32\times 32$ to facilitate the two max-pooling layers in each U-net of the DeepFDR network. 
    
    \item \textbf{fcHMRF-LIS:} Takes a 3D volume of z-statistics 
    and corresponding 3D
volume of mean-difference estimates 
 $\{\Delta\hat{\mu}_i\}_{i=1}^m$ as input.  The Python package implementing the fcHMRF-LIS method is available at  \url{https://github.com/kimtae55/fcHMRF-LIS}. 
The parameter vector $\bd{w}=(w_0,w_1,w_2)$ was initialized as $(0.5,1,1)$ by default. 
In simulation settings with weak signals
$(\mu_1\in \{0,-0.5,-1\})$, we found that initializing 
$w_2=5$ improved the spatial coherence of discoveries. Thus, we instead initialized $w_2=5$ in these cases. The non-null distribution $f_1$  was initialized
using \eqref{eq:f_1}
with $q_i(h_i=1|\bd{x})$ set to $1-$p-value$_i$. 
%    The hidden states $h_i$ are initialized using $1-$p-value, assigning higher values to stronger evidence against the null hypothesis. This initialization based on p-values serves as a convenient heuristic; if desired, users can substitute this with any prior probabilities based on domain knowledge. 
For ADNI data,
to reduce  computational burden, 50\% of voxels were randomly sampled to compute 
$\SD(\{\delta_{ij}\}_{i\ne j})$
where $\delta_{ij}\in \{\ell_{i,s}-\ell_{j,s}, \Delta\mu_i-\Delta\mu_j\}$;
The resulting standard deviations were used to define the corresponding kernel bandwidth parameters
$\{\theta_{\alpha,s}, \theta_{\gamma,s},\theta_\beta\}$.
The number of EM iterations was set to 25, with early stopping activated based on the loss $-\hat{Q}(\bd{\phi}^{(t+1)}|\bd{\phi}^{(t)})$ and a patience threshold of~5.
%, which displayed consistent convergence within 10 to 15 EM steps. 
In each EM iteration,
we minimized $-\hat{Q}_2$ in~\eqref{eq:q2} using the AdamW optimizer with an initial learning rate of $10^{-4}$ and a weight decay of 0.01, running for five epochs. 
The mean-field approximation in \eqref{gate1}-\eqref{gate2} was performed with
$R=5$ iterations each time it was applied.
%These settings remained the same for all simulations and real data analyses. However, we note that the learning rate could be varied flexibly in the $(10^{-3}, 10^{-5})$ range with no degradation in performance. 
To compute $\hat{Q}_2$ in~\eqref{eq:q2}, 
we generated
a moderate number of samples $\{h_i^{(n,t)}\}_{n=1}^{N=100}$. 
%The algorithm was tested on different sample sizes $(N=1, 50, 100, 1000)$ with no noticeable fluctuation in performance. 

\end{itemize}

In the ADNI data analysis, the original brain image volume of size 
$121 \times 145 \times 121$ was cropped to 
$100 \times 120 \times 100$ to tightly enclose the ROIs. Non-ROI voxels were set to~0 for t-statistics and z-statistics, and to~1 for p-values. Multiple testing was performed only on the ROI voxels, ensuring that non-ROI voxels were excluded from the analysis.

\bibliographystyle{elsarticle-num} 
\bibliography{main}

\end{document}